\documentclass[nonacm,sigconf]{acmart}
\pdfoutput=1
\AtBeginDocument{%
  \providecommand\BibTeX{{%
    \normalfont B\kern-0.5em{\scshape i\kern-0.25em b}\kern-0.8em\TeX}}}

\usepackage{url}
\usepackage{makecell}
\usepackage{graphicx}  
\usepackage{multirow}
\usepackage{tabularx}
\usepackage{booktabs}
\usepackage{colortbl}
\usepackage{arydshln}
\usepackage{subcaption}
\usepackage{xcolor}
\usepackage{xspace}

\newcommand{\bv}{{\mathbf v}}
\newcommand{\bw}{{\mathbf w}}

\def\eg{\emph{e.g}.\xspace}
\def\ie{\emph{i.e}.\xspace}
\def\etc{\emph{etc}.\xspace}
\def\vs{\emph{vs}.\xspace}

\definecolor{bittersweet}{rgb}{1.0, 0.44, 0.37}
\definecolor{mygreen}{rgb}{0.29, 0.7, 0.48}

\begin{document}

\title{CommerceMM: Large-Scale Commerce MultiModal Representation Learning with Omni Retrieval}

\author{Licheng Yu, Jun Chen, Animesh Sinha,
Mengjiao MJ Wang, Yu Chen, Tamara L. Berg, Ning Zhang}
\affiliation{%
  \institution{Facebook AI}
  \city{Menlo Park, California}
  \country{USA}
}
\email{{lichengyu, junchen20, animeshsinha,mengjiaow,hugochen,tlberg,ningzhang}@fb.com}

\renewcommand{\shortauthors}{Yu, et al.}

\begin{abstract}
We introduce CommerceMM - a multimodal model capable of providing a diverse and granular understanding of commerce topics associated to the given piece of content (image, text, image+text), and having the capability to generalize to a wide range of tasks, including Multimodal Categorization, Image-Text Retrieval, Query-to-Product Retrieval, Image-to-Product Retrieval, etc.
We follow the pre-training  + fine-tuning training regime and present 5 effective pre-training tasks on image-text pairs. 
To embrace more common and diverse commerce data with text-to-multimodal, image-to-multimodal, and multimodal-to-multimodal mapping, we propose another 9 novel cross-modal and cross-pair retrieval tasks, called Omni-Retrieval pre-training.
The pre-training is conducted in an efficient manner with only two forward/backward updates for the combined 14 tasks.
Extensive experiments and analysis show the effectiveness of each task.
When combining all pre-training tasks, our model achieves  state-of-the-art performance on 7 commerce-related downstream tasks after fine-tuning.
Additionally, we propose a novel approach of modality randomization to dynamically adjust our model under different efficiency constraints.
\end{abstract}



\maketitle

\section{Introduction}
\begin{figure}
    \centering
    \includegraphics[width=0.95\linewidth]{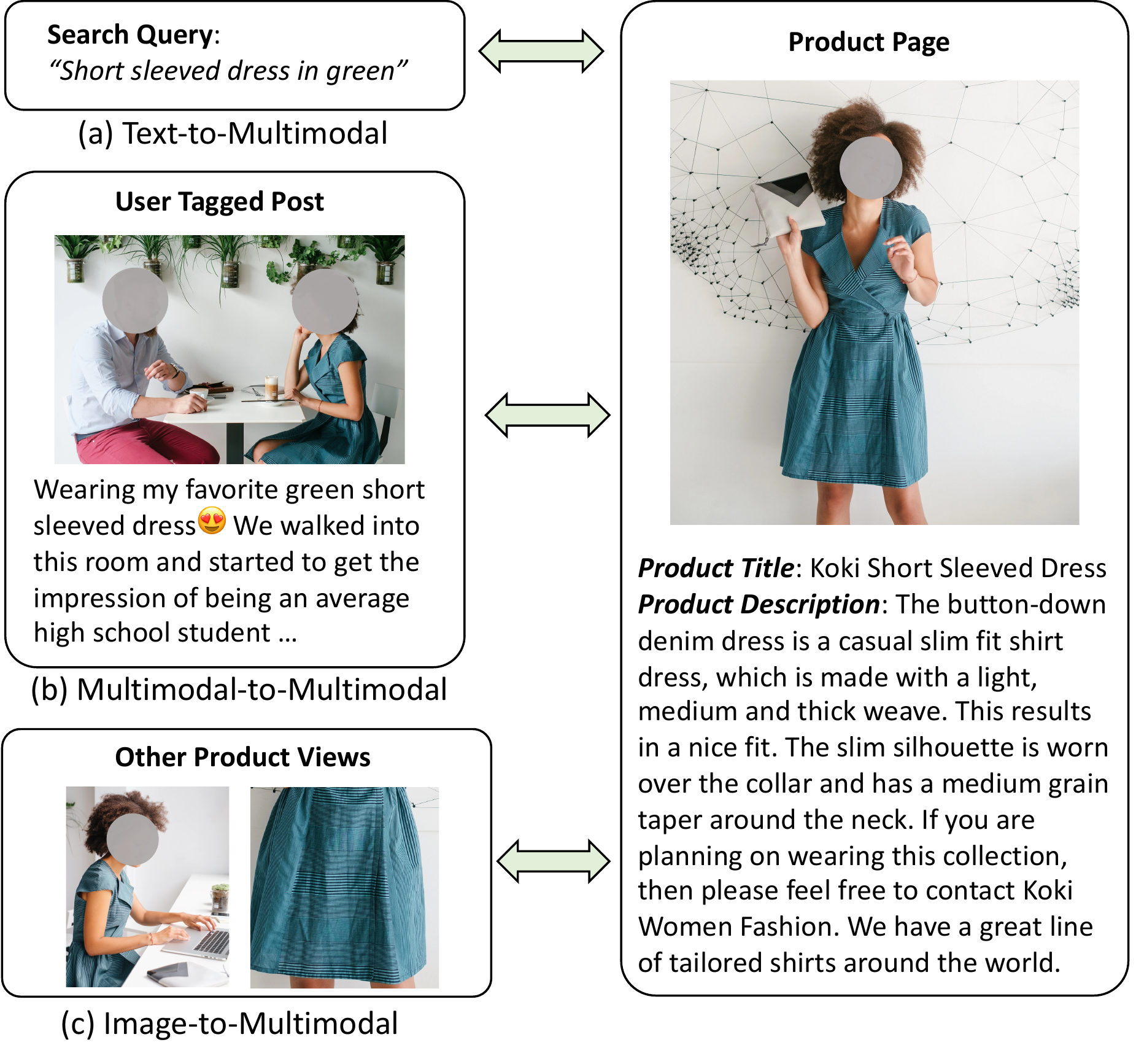}
    \caption{Example cross-modal and cross-pair data. Users could use text query to do product search. Some users tag the relevant products when uploading their multimodal media. On product page, there could be multiple views of products. While those medias are of different type (text, multimodal, image), they are linked with the same product.}
    \label{fig:example}
\end{figure}

At Facebook, nearly every post related to commerce is multimodal, e.g., a Marketplace\footnote{\url{https://www.facebook.com/marketplace}} post is composed of one or several views of a product associated with its product description, a Shop product listing is composed of the product images and detailed specifics describing the product, e.g., title, attribute, size, material, and influencers upload their fashion posts to Instagram\footnote{\url{https://www.instagram.com}} with captions and hashtags.
Since the announcement of Facebook/Instagram Shops\footnote{\url{https://www.facebook.com/business/shops}} on Facebook Business, there has been an exploding increase of usage of our commerce platforms.
For example, users perform text query search on Marketplace every second looking for specific products, users also tag the products (e.g., t-shirt, necklace) while uploading a post, they also trigger visual search looking for similar products appearing in other images on the platform, etc.
The large-scale commerce-related data and various use cases motivate us to build a commerce-specific multimodal representation for the post.

Recently vision-and-language representation learning is becoming a more and more popular research topic.
This trend has also motivated people to study the commerce-specific pre-training~\cite{gao2020fashionbert, zhuge2021kaleido, zhu2021knowledge, dong2021m5product}.
In those works, the authors pre-train the transformer-based~\cite{vaswani2017attention} model on commerce image-text pairs, then fine-tune it on image-text retrieval, image captioning, category recognition, etc.
However, most of existing works were trained on the medium-scale image-text pairs and evaluated on limited academic tasks.
This raises a question: can Facebook learn a more generalized multimodal representation for various practical commerce-related applications?

\begin{figure*}
    \centering
    \includegraphics[width=0.95\linewidth]{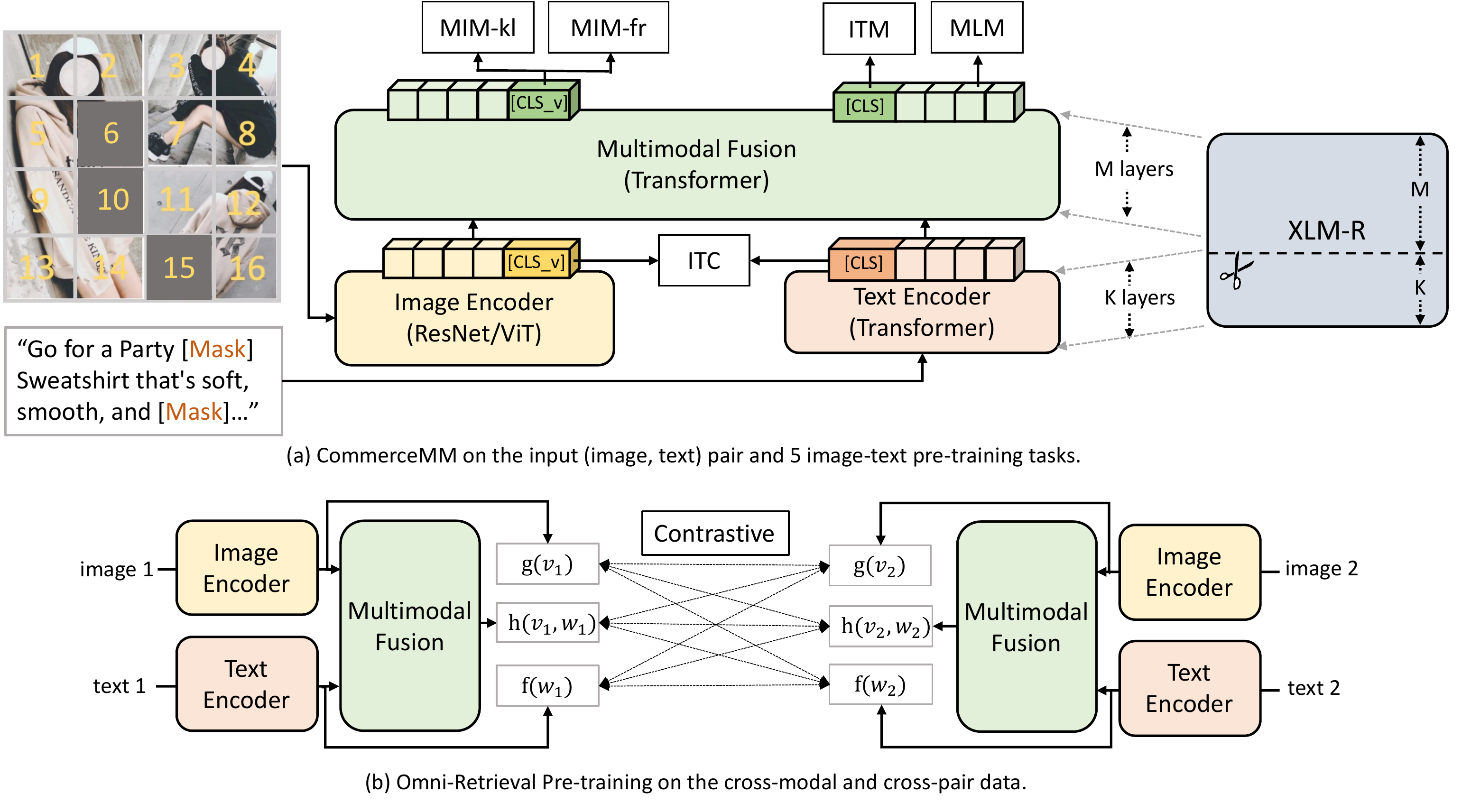}
    \caption{CommerceMM Model Architecture with the image-text pre-training and omni-retrieval tasks.}
    \label{fig:model_archi}
\end{figure*}

In this spirit, we introduce \textbf{Commerce} \textbf{M}ulti\textbf{M}odal Representation (\textbf{CommerceMM}), a large-scale pre-trained model for joint multimodal commerce embedding at Facebook.
We scale up the pre-training data from less than 1 million to hundreds of million.
Our model is composed of an image encoder, a text encoder (transformer-based), and a multimodal fusion encoder (transformer-based), as in Fig.~\ref{fig:model_archi}.
We then propose two sets of pre-training tasks for the model and evaluate on a wide range of 7 commerce-related tasks.

The first set of pre-training tasks consists of 5 effective ones on the image-text pairs, including Masked Language Modeling (MLM), Masked Image Modeling KL-Divergence (MIM-kl), Masked Image Modeling Feature Regression (MIM-fr), Image-Text Contrastive Learning (ITC), and Image-Text Matching (ITM).
While previous works~\cite{kim2021vilt, dou2021empirical} show MIM is not helpful for multimodal pre-training and typically MLM is the most effective task~\cite{li2021align, chen2019uniter}, our proposed MIM tasks are essential.
The key difference lies in ($i$) we use a larger mask ratio of 50\% instead of 15\% used in~\cite{chen2019uniter, tan2019lxmert, lu2019vilbert}; ($ii$) we only recover the signal at image's \texttt{[CLS]} token instead of reconstructing every masked patch; ($iii$) the supervision comes from the other two views (ClusterFit~\cite{yan2020clusterfit} and GrokNet~\cite{bell2020groknet}) of the intact raw image.

To our knowledge, all existing multimodal pre-training take the image-text pairs as the default input for the representation learning.
However, in practice, specifically in the commerce domain, we have access to more diverse formats of data, \eg, multiple images, texts, and other valuable data, e.g., query and click, around the same product.
For example, in Fig.~\ref{fig:example}(a), we show users could perform query search to find the related product.
User's click on some product builds a text-to-multimodal mapping between the query text and the multimodal product page.
Fig.~\ref{fig:example}(b) shows an example of users tagging the product when uploading their own multimodal media consisting of a photo and a caption, which builds a multimodal-to-multimodal mapping.
Fig.~\ref{fig:example}(c) shows a product page with multiple views of the same product, where each of them could be a source image to search for the associated product, making an image-to-multimodal mapping.
We call such data as \texttt{cross-modal and cross-pair} data.
Apparently, the common image-text pre-training tasks like ITC and ITM can no longer be applied since their input is fixed to be the image-text pair and can only handle the image-text relation.
This motivates us to propose the second set of pre-training tasks - Omni-Retrieval Pre-training.
We propose to build the relations between any modality type from image, text, and multimodality.
Specifically, we encapsulate the cross-pair data into the form of (source image, source text, target image, target text).
We predict three embeddings for the source and target respectively: visual embedding, textual embedding, and multimodal embedding.
Exhaustive matching results in 9 types of pairwise relations, where each relation can be learned with contrastive learning mimicking 9 retrieval tasks in total.
With the 9 Omni-Retrieval tasks, our pre-trained model is shown to learn a more discriminative and generalized representation for any alignment space.

We combine both 5 image-text tasks and 9 omni-retrieval tasks together during pre-training. 
While the combined pre-training (in total 14 tasks) seem unbelievably heavy at first glance, our pre-training is performed in a simple and efficient manner.
During training, there is only 1 forward/backward for each set of tasks, and we found the two sets of tasks are complementary with each other.

Besides pre-training, we conduct a preliminary exploration of another novel idea -- modality randomization.
We change the text encoder and multimodal fusion encoder layers dynamically during training.
At each training step, we randomly assign $K$ and $M$ transformer layers (while keeping the sum of $K$ and $M$ unchanged) to the text and multimodal encoder.
The shuffling role makes each layer learn from both modalities and allows knowledge sharing.
Our experiments show a light-weighted 2nd-stage pre-training with modality randomization further improves multimodal understanding.
Additionally, our model can be flexibly adjusted to different architectures per different efficiency constraints.

We introduce 7 diverse commerce tasks to evaluate the pre-trained model, including
Catalog Categorization, Marketplace Categorization, Image-to-Text Retrieval, Text-to-Image Retrieval, Query-to-Product Retrieval, Image-to-Product Retrieval, and Image-to-Image Retrieval.
We present extensive experiments and analyses to provide useful insights on the effectiveness of each pre-training task and dynamic model architecture.
The experiments show that with all 14 pre-training tasks, our model can achieve state-of-the-art performance across all  7 downstream tasks.
Furthermore, with dynamic model randomization, we can achieve an even better performance with a smaller model size.
Our contributions are summarized as follows:
($i$) We propose 5 effective image-text pre-training tasks on a large-scale commerce multimodal dataset.
($ii$) We introduce a novel pre-training approach of Omni-Retrieval for the cross-modal and cross-pair data.
($iii$) We validate our approach on 7 diverse tasks with thorough study and insightful analysis.
\section{Related Work}

Inspired by the success of BERT~\cite{devlin2018bert} in natural language processing, there has been a surge of interest in vision-and-language representation learning, by pre-training on the large-scale multimodal data with a transformer-based~\cite{vaswani2017attention} model architecture, then fine-tuning on various downstream tasks~\cite{sun2019videobert,chen2019uniter,su2019vl,lu2019vilbert,li2019visualbert,li2020unicoder,li2020hero,zhou2020unified,li2021align,wang2021simvlm,li2020unimo,zhang2021vinvl,wang2021vlmo,wang2021ufo,tan2019lxmert}.
While different works use different architectures (\eg two-stream~\cite{lu2019vilbert,tan2019lxmert,yu2020ernie} \vs single-stream~\cite{li2019visualbert,li2020unicoder,su2019vl,chen2019uniter}), features (\eg regions~\cite{anderson2018bottom} \vs grids~\cite{huang2020pixel}), backbones (\eg ConvNets~\cite{huang2020pixel} \vs Transformers~\cite{kim2021vilt}) \etc,  
the shared goal of visual-linguistic pre-training is to exploit large-scale, paired image-and-text corpora~\cite{lin2014microsoft,krishna2017visual,sharma2018conceptual,ordonez2011im2text,jia2021scaling,radford2021learning} and obtain models that are \emph{implicitly} pre-built with the appropriate correspondences to understand the multi-modal data. 
The level of understanding is benchmarked with multiple multimodal downstream tasks~\cite{zellers2019from,plummer2015flickr30k,yu2016modeling,antol2015vqa}.

Various tasks have been introduced for the multi-modal pre-training. 
The most notable three are Masked Language Modeling (MLM), Masked Image Modeling (MIM), and Image-Text Matching (ITM) -- which are the direct counterparts of the BERT objectives. 
Several other variants of these three tasks have also been explored, such as predicting object tags~\cite{li2020oscar,hu2020vivo}, masked region classification~\cite{lu2019vilbert,chen2019uniter}, sequential caption generation~\cite{zhou2020unified,wang2021simvlm}, image-text contrastive learning~\cite{li2021align,jia2021scaling,radford2021learning}, etc.
Recent works~\cite{dou2021empirical,kim2021vilt} found MIM may not be essential for pre-training among those tasks.
In this work, we found that our proposed MIM is more effective than MLM.
Inspired by recent works of self-supervised learning on vision~\cite{xie2021simmim,he2021masked}, we propose to mask out image patches with larger proportion and follow MaskFeat~\cite{wei2021masked} to reconstruct other views of the whole image rather than recovering those masked regions only.

Besides, all existing pre-training works are conducted on the image-text pairs.
We can certainly convert the cross-modal and cross-pair data in Fig.~\ref{fig:example} to the image-text pairs, but that would constraint the model from learning more general representations in other cross-modal space, e.g., multimodal-to-multimodal alignment.
In comparison, we propose a novel approach of Omni-Retrieval pre-training to build the relations between any modality towards learning a universal representation.

\section{Model and Pre-training}

\subsection{Model Overview}\label{sec:model_overview}
The model architecture of CommerceMM is illustrated in Fig.~\ref{fig:model_archi}(a), which is composed of an image encoder, text encoder, and a multimodal fusion encoder.
Given an image, the vision encoder converts the raw pixels into a sequence of visual embeddings, \ie, flattened grid/patch-wise features.
On the other side, the text encoder converts the input sentence into a sequence of word tokens and feeds them into the transformer to get textual embeddings.
Inside both visual and textural embeddings, there is a special \texttt{[CLS]} token, which encodes the whole image and sentence representation.
We feed both sequences of embeddings into another transformer - multimodal fusion encoder, where the two modalities learn to interact with each other building a contextualized multimodal embeddings.
We still keep the \texttt{[CLS]} tokens of both modalities at the output of multimodal fusion encoder.

Our image encoder is an off-the-shelf ResNet or ViT-B/16 model, that have been weakly-supervised trained on billions of Instagram images using hashtags~\cite{mahajan2018exploring}.
Our text encoder and multimodal fusion encoder are initialized from XLM-R~\cite{conneau2019unsupervised}, an unsupervised multi-lingual language model.
Specifically, the text encoder inherits its first $K$ layers and the multimodal fusion encoder inherits its remaining $M$ layers, whose sum equals to XLM-R's total layers (6, 12, or 24 depending on the chosen model size).
We keep $K$ and $M$ unchanged under most setting for a fixed model architecture.

Interestingly, such design also allows us to dynamically change $K$ and $M$ during training, \ie, each layer shuffles between text encoder and multimodal encoder.
In Sec.~\ref{sec:random_k}, we show the modality randomization can further enhance the multimodal understanding and make the model adjustable per different efficiency constraints.

Given an image-text pair, we denote its raw visual inputs as $\bv_i = \{\bv_i\textsuperscript{cls}, \bv_i^1, ..., \bv_i^K\}$,
and its input words as $\bw_i = \{\bw_i\textsuperscript{cls}, \bw_i^1, ..., \bw_i^T \}$,
where the subscript $i$ indicates the $i$-th pair in the dataset.
As above, an additional special \texttt{[CLS]} token is inserted to each sequence.

We introduce two sets of pre-training tasks: ($i$) Image-Text Pre-training, which consists of 5 tasks of MLM, MIM-kl, MIM-fr, ITC, and ITM as Fig.~\ref{fig:model_archi}(a); ($ii$) Omni-Retrieval Pre-training, which consists of 9 cross-modal and cross-pair retrieval tasks as Fig.~\ref{fig:model_archi}(b).
We denote $\mathbf{m}$ as the mask indices, which is applied to mask out part of the sequences for self-supervised learning.
The details of each task are presented as follows.

\subsection{Image-Text Pre-training}\label{sec:img_txt_pretrain}
\noindent \textbf{Masked Language Modeling (MLM)}
In MLM, we randomly mask out the input words with probability of 15\% and replace the masked ones $\mathbf{w}^\mathbf{m}$ with special token \texttt{[MASK]}.
The goal of MLM is to predict these masked words based on their surrounding words $\mathbf{w}^{\setminus \mathbf{m}}$ and the visual context $\mathbf{v}$, by minimizing the negative log-likelihood:
\begin{equation}\nonumber
    \mathcal{L}_{\text{MLM}}(\theta) = -\mathbb{E}_{(\mathbf{w}, \mathbf{v})\sim D} \log P_{\theta} ( \mathbf{w}^\mathbf{m} | \mathbf{w}^{\setminus \mathbf{m}}, \mathbf{v} ),
\end{equation}
where $\theta$ is the model parameters.
Each pair $(\mathbf{w}_i, \mathbf{v}_i)$ is sampled from the whole training set $D$.

\noindent \textbf{Masked Image Modeling (MIM)}
Similar to MLM, we also sample and mask the visual inputs, e.g., patches.
Previous works~\cite{kim2021vilt, dou2021empirical} found the naive masked patch regression is not helpful in multimodal pre-training.
In this work, we disregard the reconstruction of each masked region, but instead recover the holistic image signal at $\bv\textsuperscript{cls}$ token.
We first follow~\cite{xie2021simmim, he2021masked} to use a larger masking ratio of 50\% (instead of 15\% as in~\cite{chen2019uniter, tan2019lxmert, lu2019vilbert}).
The masked patches are replaced with grey pixels.
Our supervision is provided by another view of the original intact input image. 
While MaskFeat~\cite{wei2021masked} used Histograms
of Oriented Gradients (HOG) as the supervision for visual pre-training,
we rely on the more discriminative signals from ClusterFit~\cite{yan2020clusterfit} and GrokNet~\cite{ bell2020groknet} to extract additional two views of the raw image.
Between the two,~\cite{yan2020clusterfit} provides the clustering probability $c(\mathbf{v})$ while~\cite{bell2020groknet} extracts the pool5 embedding $r(\mathbf{v})$ (feature output from the 5-th Conv Block).
Correspondingly, we propose two MIM tasks but sharing the same objective base:
\begin{equation}\nonumber
    \mathcal{L}_{\text{MIM}}(\theta) = \mathbb{E}_{(\mathbf{w}, \mathbf{v})\sim D} d_{\theta}(\mathbf{v}^\mathbf{m} | \mathbf{v}^{\setminus \mathbf{m}}, \mathbf{w}),
\end{equation}
where $d_{\theta}$ is defined as follows.

(1) \textbf{Masked Image Modeling Feature Regression (MIM-fr)}
MIM-fr learns to regress the multimodal transformer's output at $\bv_i\textsuperscript{cls}$ to the pool5 embedding from GrokNet, \emph{i.e.}, $r(\mathbf{v})$.
Specifically, we apply an FC layer to convert its hidden output into a vector $\phi_{\theta}(\bv_i\textsuperscript{cls})$ of the same dimension as GrokNet's pool5.
Then we apply L2 regression on the mean square error between the two: 
$d_\theta (\bv\textsuperscript{cls}) | \bv^{\setminus \mathbf{m}}, \bw) = \| \phi_{\theta}(\bv\textsuperscript{cls}) -  r(\mathbf{v}) \|_2^2$.

(2) \textbf{Masked Image Modeling KL-Divergence (MIM-kl)}
MIM-kl applies the soft label of ClusterFit probability\cite{yan2020clusterfit} as the supervision signal, which is the softmax from ClusterFit output formatting the distribution of clusters $\tilde{c}(\mathbf{v})$.
We project the multimodal transformer's output at $\bv_i\textsuperscript{cls}$ to the same distribution space $\psi_\theta(\bv_i\textsuperscript{cls})$ with softmax.
We aim to distill the intact knowledge from ClusterFIT into CommerceMM, by minimizing the KL divergence between the two distributions:
$
d_\theta (\bv\textsuperscript{cls} | \bv^{\setminus \mathbf{m}, \bw} ) = \| 
D_{KL}( \tilde{c}(\mathbf{v}) || \psi_{\theta}(\bv\textsuperscript{cls})
\|
$.

\noindent \textbf{Image-Text Contrastive Learning (ITC)}
Following~\cite{li2021align}, we add an image-text contrastive loss between the visual and textual embeddings right before feeding them into the multimodal fusion module.
It aims to align the two modalities into the same space before fusion.
Specifically, we project the textual embedding $\bw_i\textsuperscript{cls}$ and visual embedding at $\bv_i\textsuperscript{cls}$ to a normalized lower-dimensional representations via two linear transformations $f$ and $g$.
The similarity of the text $\bw_i$ and the image $\bv_j$ is then measured by the dot product of
\begin{equation}\nonumber
 s(\bw_i, \bv_j) = f_{\theta}(\bw_i\textsuperscript{cls})^T g_{\theta}(\bv_i\textsuperscript{cls})
\end{equation}
We apply contrastive learning to bring the matched image-text pairs in the embedding space closer than the unmatched ones as follows:
\begin{align}
\mathcal{L}_{itc} = \nonumber
-\sum_i[\log\frac{\exp( s(\bw_i, \bv_i) / \tau ) }{ \sum_j \exp( s(\bw_i, \bv_j) / \tau ) } 
+ 
\log\frac{\exp( s(\bw_i, \bv_i) / \tau ) }{ \sum_j \exp( s(\bw_j, \bv_i) / \tau ) }
],
\end{align}
where $\tau$ is a learned temperature parameter.

\noindent \textbf{Image-Text Matching (ITM)}
In ITM, the inputs are a paired sentence and image and the output is a binary label $y\in \{0, 1\}$, indicating if each input pair is a match.
We extract the hidden output of $\bw\textsuperscript{cls}$ at the last layer of multimodal fusion to represent the fused representation of both modalities, then feed it into an FC layer followed by a sigmoid function to predict a single score between 0 and 1.
We denote the output score as $o_{\theta}(\mathbf{w}, \mathbf{v})$.
During pre-training, we sample a positive or negative pair $(\mathbf{w}, \mathbf{v})$ from the dataset $D$ at each step.
The negative pair is created by replacing the image or text in a paired sample with a randomly-selected one from other samples.
Following~\cite{li2021align}, when ITC is applied, we could sample the hard negative pairs from ITC's computed similarity matrix $s(\bw_i, \bv_j)$.
The incorporation of those hard negatives makes ITM a harder task, which is more beneficial for the pre-training~\cite{miech2021thinking}.
We apply binary cross entropy for this loss:
\begin{equation}\nonumber
    \mathcal{L}_{\text{ITM}}(\theta) = - \mathbb{E}_{(\mathbf{w}, \mathbf{v})\sim D} [y \log o_{\theta}(\mathbf{w}, \mathbf{v}) + (1-y) \log (1-o_{\theta}(\mathbf{w}, \mathbf{v}))] ).
\end{equation}

\subsection{Cross-Pair Pre-training: Omni Retrieval}
As in Fig.~\ref{fig:example}, besides image-text pairs, there are also a huge amount of cross-modal and cross-pair commerce data.
We formulate such data as two pairs, where we denote the source pair as $(\bw_i, \bv_i)$ and the target pair as $(\bw_j, \bv_j)$. 
Note one of $\bw$ and $\bv$ (in the source/target pair) could be missing in some case.
For example, the source for search query is only one single sentence, and the source for visual search is only one single image, while both are linked with some multimodal product pages.
We replace the missing image or text with grey pixels or an empty string and introduce an indicator $\delta$ to tell the existence of each modality.
We replicate our model for both source and target pairs sharing the same parameters, as in Fig.~\ref{fig:model_archi}(b).
We first feed the source pair $(\bw_i, \bv_i)$ to our model, our image encoder, text encoder, and multimodal fusion return three embeddings at their corresponding \texttt{[CLS]} respectively.
With 3 simple linear transformations, we get three normalized embeddings $g(\bv_i)$, $f(\bw_i)$, and $h(\bw_i, \bv_i)$.
Similarly, we can get the image embedding $g(\bv_j)$, text embedding $f(\bw_j)$, and multimodal embedding $h(\bw_j, \bv_j)$ for the target pair.
If a source pair is linked with a target pair, we assume any existing modality from the source would be highly correlated with every existing modality from the target.
Thus we compute the similarity score between any pair of source and target embedding from the text, image, or multimodal perspective respectively as the follows:
\begin{align*}\nonumber
\small
\begin{gathered}
  \text{Text:} 
  \begin{cases} 
    s^{T\leftrightarrow  T}_{ij} = f(\bw_i)^T f(\bw_j)  \\
    s^{T\leftrightarrow  I}_{ij} = f(\bw_i)^T g(\bv_j)  \\
    s^{T\leftrightarrow  M}_{ij} = f(\bw_i)^T h(\bw_j, \bv_j),
  \end{cases} 
  \text{Image:}
  \begin{cases} 
    s^{I\leftrightarrow  T}_{ij} = g(\bv_i)^T f(\bw_j)  \\
    s^{I\leftrightarrow  I}_{ij} = g(\bv_i)^T g(\bv_j)  \\
    s^{I\leftrightarrow  M}_{ij} = g(\bv_i)^T h(\bw_j, \bv_j),
  \end{cases}
  \\\nonumber
  \text{Multimodal:}
  \begin{cases} 
    s^{M \leftrightarrow T}_{ij} = h(\bw_i, \bv_i)^T f(\bw_j) \\
    s^{M \leftrightarrow I}_{ij} = h(\bw_i, \bv_i)^T g(\bv_j) \\
    s^{M \leftrightarrow M}_{ij} = h(\bw_i, \bv_i)^T h(\bw_j, \bv_j).
  \end{cases}
\end{gathered}
\end{align*}
In total, there are 9 cross-modal combinations, resulting in 9 similarity matrices within each batch.
We define our Omni-Retrieval loss as the sum of the contrastive loss over the 9 similarities:
\begin{align} \nonumber
    \mathcal{L}_{omni} = \sum_{u, v \in \{I, T, M\}}\sum_i \delta^{u}_i [ &\log \frac{ \exp(s^{u \leftrightarrow v}_{ii}/\tau) }{ \sum_j \exp(s^{u \leftrightarrow  v}_{ij}/\tau) }  \\\nonumber
    + &\log \frac{ \exp(s^{u \leftrightarrow v}_{ii}/\tau) }{ \sum_j \exp(s^{u \leftrightarrow  v}_{ji}/\tau) } ],
\end{align}
where $u$ and $v$ refers to the modality of image, text, or multimodal, and $\delta^{u}_i$ indicates if the modality $u$ of the $i$-th input pair exists or not, which works like a gate function to turn on/off the contrastive learning for each pair.
\section{Experiments}
We report experimental results on three model sizes: ResNet50-based 6-layer CommerceMM, ViT-B/16-based 6-layer CommerceMM, and ViT-B/16-based 12-layer CommerceMM\footnote{
Our ResNet50-based 6-layer CommerceMM (L=6, H=512, A=8) has 172M parameters; ViT-B/16-based 6-layer CommerceMM has 234M parameters; 
ViT-B/16-based 12-layer-CommerceMM (L=12, H=768, A=12) has 365M parameters. 
(L: total number of transformer layers (text + multimodal fusion encoders); 
H: hidden activation dimension; 
A: number of attention heads. 
Inside each model, there are 128M parameters from XLM-R's token embeddings.)}.
We use MMF~\cite{singh2020mmf} for the implementation.

During pre-training, we set the learning rate as 5e-5, batch size as 2,048, and update in total 300K steps.
We apply another 100K steps for a 2nd-stage pre-training when modality randomization in Sec.~\ref{sec:random_k} is applied.
In the 5 image-text pre-training tasks, we empirically assign an equal weight of 1.0 to MIM-kl, MIM-fr, ITM and ITC, and 0.5 to MLM.
For the 9 Omni-Retrieval tasks, we also assign an equal weight of 1.0 to each of them.
While there are in total 14 different tasks, our model is trained efficiently in a round-robin fashion with only two types of forward/backward.
At each step, we randomly pick between the set of 5 image-text pre-training tasks or the set of 9 Omni-Retrieval tasks for updating the model parameters.
It takes in total 3,840 and 4,608 A100 hours to pre-train our end-to-end 6-layer CommerceMM and 12-layer CommerceMM respectively.
We evaluate the pre-trained models on 7 commerce-related tasks.

\subsection{Pre-training Dataset}
Our pre-training dataset consists of two subsets.
The first subset is 102M image-text pairs, collected from two sources -- Product Catalog Posts (50M) and Marketplace Posts (52M). 
The text is the concatenation of product title and description. 
The average number of words of the concatenated texts is 62. 
Note our texts are multi-lingual, including English, French, German, Spanish, Portuguese and other languages.
For each image, we pre-compute its GrokNet embeddings~\cite{ bell2020groknet} and ClusterFit probability~\cite{yan2020clusterfit} for the MIM tasks in our image-text pre-training.

The second subset is 50M cross-modal and cross-pair data for the Omni-Retrieval pre-training.
The data comes from various sources: 
\begin{enumerate}
    \item IG and FB Shops product catalog: Each product has product title/description and product image.
    \item IG and FB Shops text search queries with clicked product: The search query text is used as the query to retrieve the clicked products (product title/description, product image).
    \item IG and FB posts where a product is tagged on the post: The post usually contains a post image and post caption text. The post caption text is frequently empty. Such posts are used to retrieve the tagged product (product title/description, product image) given the post image.
\end{enumerate}

\subsection{Downstream Tasks}
With minimal surgery of CommerceMM, our model can be adapted to new tasks that are not seen during pre-training.
We introduce 7 downstream commerce-related tasks.
Each of the downstream fine-tuning tasks has its own dataset and annotations.

\noindent\textbf{Catalog Categorization (CC)}
We annotated the fine-grained category labels for 2.5M multimodal catalog posts.
The labels are tree-structured on shopping-based categories, e.g., ``Home/Home Decor/Decorative Accents/Posters, Prints \& Paintings".
There are in total 5,168 leaf labels.
We focus on classifying the leaf label for each post.
One example of our catalog posts is shown in the product page of Fig.~\ref{fig:example}, consisting of an image, a title and a description\footnote{For those posts with multiple images, we select the first one to pair with the description making an image-text pair.}.
We concatenate the title and description as a full sentence.
There are on average 99.6 words for each post.
We add a 5,168-way classifier on top of the $\bw\textsuperscript{cls}$ of the multimodal fusion encoder and apply cross entropy loss during the fine-tuning.
Overall accuracy is reported as the evaluation metric.

\begin{table*}[t!]
\centering
\resizebox{2.05\columnwidth}{!}{
\begin{tabular}{ p{0.02\textwidth} l | c | c  c c c c c c}
\hline
 & Pre-training Tasks & Meta Avg. & CC & MPC & T2I & I2T & Q2P & I2P & I2P\textsuperscript{i} \\ 
\hline
 \small{1} & \small{None} & 50.39 & 72.08 & 63.75 & 22.70 & 23.88 & 48.43 & 55.80 & 66.10 \\
 \small{2} & \small{MLM} & 52.77 & 73.10 & 67.94 & 24.87 & 25.84 & 51.59 & 60.46 & 65.59 \\
 \small{3} & \small{MIM-kl} & 53.59 & 73.26 & 69.04 & 26.31 & 26.91 & 53.89 & 59.18 & 66.54 \\
 \small{4} & \small{MIM-kl + MIM-fr} & 54.18 & 73.27 & 69.12 & 27.88 & 28.61 & 54.05 & 59.48  & 66.83 \\
 \small{5} & \small{MLM + MIM-kl + MIM-fr} & 54.19 & 73.64 & 69.55 & 26.66 & 26.98 & 53.47 & 61.66 & 67.30 \\

 \small{6} & \small{MLM + MIM-kl + MIM-fr + ITM} & 54.62 & 73.64 & 69.45 & 27.82 & 28.33 & 54.63 & 61.63 & 66.87 \\
 \small{7} & \small{MLM + MIM-kl + MIM-fr + ITM + ITC} & \cellcolor[gray]{.8}57.87 & \cellcolor[gray]{.8}73.76 & \cellcolor[gray]{.8}69.61 & \cellcolor[gray]{.8}39.11 & \cellcolor[gray]{.8}40.30 & 55.60 & 60.03 & 66.65 \\
 \small{8} & \small{Omni Retrieval (Omni)} & 56.27 & 72.98 & 67.81 & 29.69 & 30.78 & \cellcolor[gray]{.8}57.34 & \cellcolor[gray]{.8}67.98 & \cellcolor[gray]{.8}67.31 \\
 \small{9} & \small{MLM + MIM-kl + MIM-fr + ITM + ITC + Omni} & \cellcolor[gray]{.6}60.64 & \cellcolor[gray]{.6}73.77 & \cellcolor[gray]{.6}69.73 & \cellcolor[gray]{.6} 42.05 & \cellcolor[gray]{.6}43.06 & \cellcolor[gray]{.6}58.48 & \cellcolor[gray]{.6}69.20 & \cellcolor[gray]{.6}68.16 \\
\hline
\end{tabular}
}
\vspace{1pt}
\caption{Ablation Study of different pre-training tasks using the fine-tuned performance of Catalog Categorization (CC), Marketplace Categorization (MPC), Text-to-Image Retrieval (T2I), Image-to-Text Retrieval (I2T), Query-to-Product Retrieval (Q2P), Image-to-Product Retrieval (I2P), and Image-to-Product-Image Retrieval (I2P\textsuperscript{i}). 
All results are obtained from ResNet50-based CommerceMM (6-layer).
For all retrieval tasks, R@1 scores are reported. 
Meta Avg is the average score of 7 downstream tasks, measuring the overall performance.
Dark and light grey colors highlight the top and second best results for each task.}
\label{table:ablation_study}
\end{table*}

\noindent\textbf{Marketplace Categorization (MPC)}
Similar to Catalog Categorization, we have also annotated the fine-grained category labels for 2M Marketplace posts.
There are in total 1,387 leaf labels for the collected Marketplace posts.
Each post consists of an image, a title and a description. 
We follow the same process concatenating the title and description.
On average, there are 37.3 words in each text.
We add a 1,387-way classifier on the multimodal fusion encoder's  $\bw\textsuperscript{cls}$ token for fine-tuning and inference.
Overall accuracy is reported as the evaluation metric.

\noindent\textbf{Image-Text Retrieval}  
We collect 1M Catalog Posts for the text-to-image retrieval (\textbf{T2I}) and the image-to-text retrieval (\textbf{I2T}) tasks.
We split a 10K subset for evaluation and leave the rest for model training.
There are two ways to perform the image-text retrieval.
One is using the ITM head to predict the matching score between the input image-text pair and rank the scores of all pairs~\cite{chen2019uniter, gao2020fashionbert, zhuge2021kaleido}. 
One is computing the similarity between image encoder and text encoder and picks the best match without deeper fusion between the two modalities as~\cite{jia2021scaling,radford2021learning,li2021align}.
While the first approach may achieve better performance~\cite{miech2021thinking}, its computational cost is huge (quadratically increasing w.r.t the size of retrieval pool), which is not applicable in practice. 
Thus we follow the second approach to simply use our image and text encoders of CommerceMM for the retrieval tasks.
We apply contrastive learning loss during the fine-tuning.
Recall@1 is used to measure the retrieval performance.

\begin{figure}
    \centering
    \includegraphics[width=0.95\linewidth]{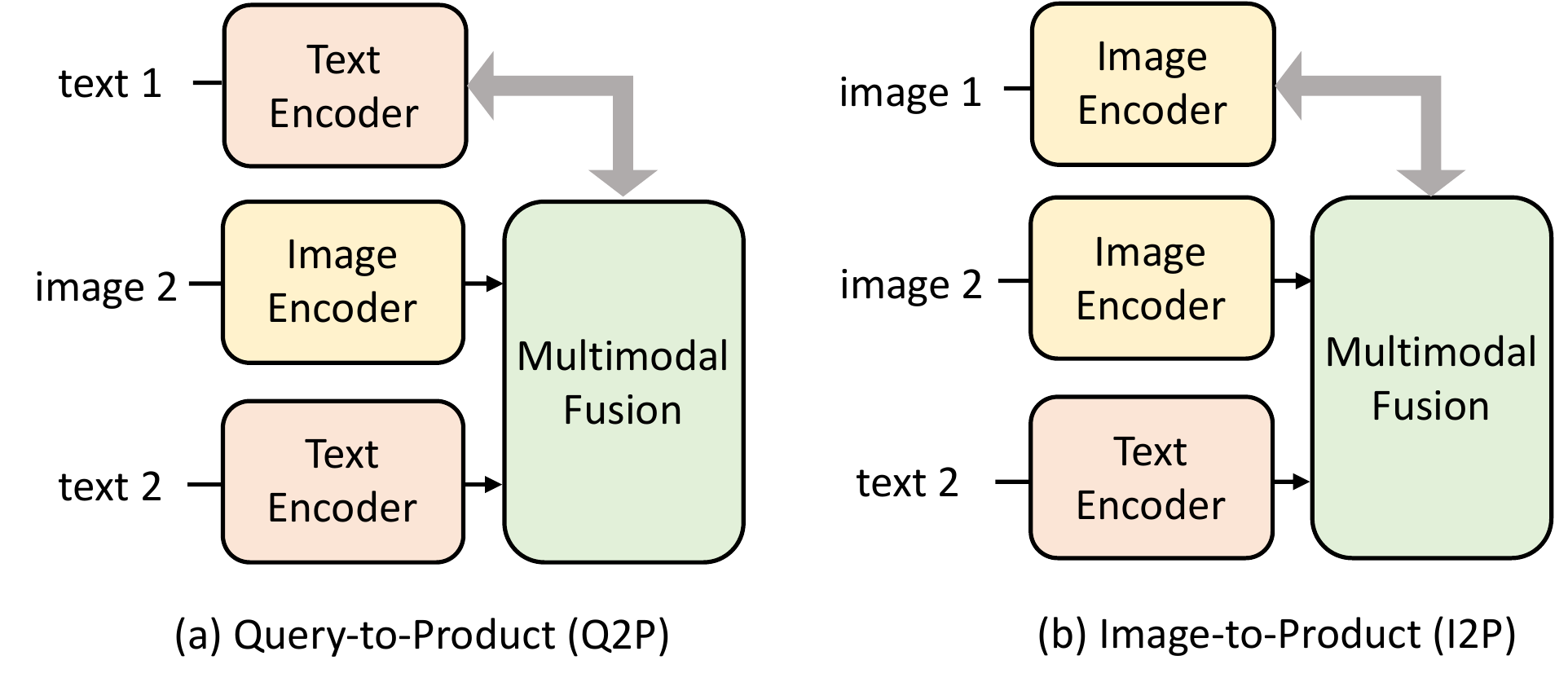}
    \caption{The model architecture for the Query-to-Product and Image-to-Product retrieval tasks.}
    \label{fig:downstream_model}
\end{figure}

\noindent\textbf{Query-to-Product Retrieval (Q2P)} 
We collected 1M text search queries with their clicked products.
Each product is associated with a multimodal product page.
We split a 10K subset for evaluation and leave the rest for model training.
We consider this problem as a text-to-multimodal retrieval task. 
The model is shown in Fig.~\ref{fig:downstream_model}(a), where we use the CommerceMM's text encoder to encode the input query text and whole model to encode the candidate multimodal post.
Contrastive learning loss is applied during fine-tuning.
We use Recall@1 as the evaluation metric.

\noindent\textbf{Image-Product Retrieval} 
We collected 1M post image queries with the tagged products.
Each pair consists of a source image and a target product page.
We split out a 10K subset for evaluation and set up two tasks for this dataset.
The first is the image-to-multimodal retrieval as the given data format, which is called image-to-product retrieval (\textbf{I2P}).
The model for I2P is shown in Fig.~\ref{fig:downstream_model}(b), where we use the image encoder to encode the input query image and the whole model to encode the multimodal product post.
The second is the image-to-image retrieval task, where we only use the target image as the candidates. 
We call this task as image-to-product-image retrieval (\textbf{I2P\textsuperscript{i}}).
For both tasks, we apply contrastive learning loss and use Recall@1 for evaluation.


\subsection{Ablation Study on Pre-training}\label{sec:ablation}
We analyze the effectiveness of different pre-training settings through ablation studies over the 7 downstream tasks.
In addition to the above mentioned standard metrics, we also compute the Meta Average score (average of the results across all tasks) as a global metric.
The complete ablation study is listed in Table.~\ref{table:ablation_study}.
In these experiments, we use a ResNet50-based 6-layer CommerceMM, \ie, the number of layers of text encoder and multimodal fusion together equals 6, initialized from XLM-R-small~\cite{conneau2019unsupervised}.
All models are trained in an end-to-end manner.

First, we provide a baseline without any multimodal pre-training involved in Line 1 (L1).
In other words, this model is directly initialized from the off-the-shelf ResNet and XLM-R directly, which were pre-trained in vision-only and language-only domain.

Second, we validate the effectiveness of each pre-training task through a thorough ablation.
Comparing L2 and L3, our proposed MIM-kl (L3) achieves a clear gain over MLM (L2).
When applying both MIM-kl and MIM-fr, L4 further improves the performance.
This is a quite different observation compared with~\cite{kim2021vilt, dou2021empirical}, where previously MIM was not shown to be helpful.
The difference indicates the effectiveness of our proposed MIM tasks.
Interestingly, the combined MLM, MIM-kl, and MIM-fr in L5 does not quite outperform L4 (with only 0.01 gain on the Meta Average).
One possible reason might be the two MIM tasks overshadows the effect of MLM.
L6 and L7 adds ITM and ITC into the pre-training, both of which further improves the meta average.
Notably, ITC introduces a significant improvement on image-text retrieval tasks, \ie, T2I and I2T due to the same task has been well warmed up during pre-training.

Next, we validate the contribution of Omni-Retrieval (Omni) pre-training tasks.
In L8, we only apply the 9 Omni-Retrieval tasks during pre-training without any help from image-text pre-training.
We observe a significant gain on Q2P, I2P, and I2P\textsuperscript{i}, which benefits from the text-to-multimodal, image-to-multimodal, and image-to-image tasks in the Omni.
Additionally, Omni also helps on the first 4 image-text tasks (CC, MPC, T2I, I2T), comparing L1 and L8.
When combining both image-text and Omni-Retrieval pre-training, our model in L9 achieves the best across every single task. 

Last but not least, we compare the performance between I2P and I2P\textsuperscript{i}.
The two tasks share the same evaluation set, where each image in I2P\textsuperscript{i} is from its corresponding product page in I2P.
Interestingly, we found that without Omni Retrieval, I2P\textsuperscript{i} always performs better than I2P in L1-L7. 
As comparison, L8 and L9 shows I2P's results are better than I2P\textsuperscript{i} with Omni, which aligns with our intuition that the multimodal product page contains more cues than its product image only.
This indicates Omni helps learning more generalized representations under different alignment space.
\begin{table*}[t]
\centering
\resizebox{2.1\columnwidth}{!}{
\begin{tabular}{c l c l | c |  c c c c c c c }
\hline
& Vis. Enc. & (K, M) & Pre-training Tasks & Meta Avg. & CC & MPC & T2I & I2T & Q2P & I2P & I2P\textsuperscript{i} \\ 
\hline
 1 & GrokNet Hash & (0, 6) & None & - & 71.26 & 65.26 & - & - & - & - & - \\
 2 & GrokNet Hash & (0, 6) & \small{MLM + MIM-kl + MIM-fr + ITM} & - & 73.40 & 69.04 & - & - & - & - & - \\
 3 & ResNet50 & (0, 6) & \small{MLM + MIM-kl + MIM-fr + ITM} & - & 73.54 & 69.13 & - & - & - & - & - \\
 4 & ResNet50 & (3, 3) & \small{MLM + MIM-kl + MIM-fr + ITM} & 54.62 & 73.64 & 69.45 & 27.82 & 28.33 & 54.63 & 61.63 & 66.87 \\
 5 & ResNet50 & (3, 3) & \small{MLM + MIM-kl + MIM-fr + ITM + ITC + Omni} & 60.64 & 73.77 & 69.73 & 42.05 & 43.06 & 58.48 & 69.20 & 68.16 \\
 6 & ViT-B/16 & (3, 3) & \small{MLM + MIM-kl + MIM-fr + ITM + ITC + Omni} & \cellcolor[gray]{.8}62.69  & \cellcolor[gray]{.8}73.78 & \cellcolor[gray]{.8}69.80 & \cellcolor[gray]{.8}43.84 & \cellcolor[gray]{.8}44.29 & \cellcolor[gray]{.8}61.43 & \cellcolor[gray]{.8}73.41 & \cellcolor[gray]{.8}72.31 \\
 7 & ViT-B/16 & (6, 6) & \small{MLM + MIM-kl + MIM-fr + ITM + ITC + Omni} & \cellcolor[gray]{.6}66.85 & \cellcolor[gray]{.6}74.31 & \cellcolor[gray]{.6}70.60 & \cellcolor[gray]{.6}52.10 & \cellcolor[gray]{.6}53.72 & \cellcolor[gray]{.6}65.70 & \cellcolor[gray]{.6}77.16 & \cellcolor[gray]{.6}74.36 \\
\hline
\end{tabular}
}
\vspace{1pt}
\caption{Effect of vision encoder, text encoder, and model size. (K, M) stands for (\#text layers, \#multimodal layers) inside CommerceMM, both of which are transformers. Dark and light grey colors highlight the top and second best results.
}
\label{tab:model_archi}
\end{table*}

\subsection{Effectiveness of Vision Encoder, Text Encoder, and Model Size}
\label{sec:exp_effect}
In Table~\ref{tab:model_archi}, we first show the results using the off-the-shelf GrokNet~\cite{bell2020groknet} Hash feature as the image embedding.
We feed the text tokens and image embedding directly into the multimodal transformer as~\cite{chen2019uniter,li2019visualbert}, \ie, the multimodal model is an early-fusion model without text encoder.
We apply MLM, MIM-kl, MIM-fr and ITM for the multimodal pre-training, each of which is same as in Sec.~\ref{sec:img_txt_pretrain}\footnote{We cannot apply image-text contrastive learning or omni retrieval as there is no text encoder in the early-fusion architecture.}.
The mere difference lies in the masking strategy of MIM, where we randomly shuffle the 0/1 bits of the input Hash for masking. 
For fine-tuning, we focused on the multimodal categorization tasks of CC and MPC.
We observe that even with only 4 pre-training tasks, there is a significant improvement on CC and MPC tasks from L1 over L2, indicating that our pre-training also works well with a fixed vision encoder of GrokNet.
We launched L2 as our current production model. 
More details of the deployment and product impact are provided in Sec.~\ref{sec:deployment}.
We then show the advantage of end-to-end training.
Comparing L2 and L3, we observe the end-to-end trained model with ResNet50 already outperforms GrokNet's Hash~\cite{bell2020groknet}, which is from a ResNeXt101 model.

Next, we compare with and without text encoder.
Both L3 and L4 have the same number of transformer layers, \ie, the model size is the same, and are pre-trained with the same tasks.
We observe that with a 3-layer text encoder, the model in L4 achieves better performance than the early-fusion model in L3 on CC and MPC.
Moreover, the introduction of a text encoder allows us to perform those text-based retrieval tasks, \eg, T2I, I2T, and Q2P.
Adding the Omni-Retrieval pre-training tasks, the model in L5 further improves the performance of all the retrieval tasks, compared with L4.

We then compare the effectiveness of vision encoder.
With the same input image size 224x224 and the same transformer architecture, we found ViT-B/16~\cite{dosovitskiy2020image} (L6) brings a consistent gain over ResNet50 (L5) on each of the 7 downstream tasks, showing a better visual encoder is beneficial to the multimodal applications.

We also experiment scaling up our transformer from 6 layers (L6) to 12 layers (L7).
Note for the 12-layer transformer, we simply assign its first 6 layers from XLM-R to the text encoder and leave the rest 6 layers to the multimodal fusion encoder.
Comparing L6 and L7, we observe a further improvement on each task with the larger model.
We leave the exploration of even larger transformer with more advanced vision encoder to the future work.

\subsection{Transferability to Academic Dataset}

We also evaluate how our pre-trained model performs on the academic dataset, \eg, FashionGen~\cite{rostamzadeh2018fashion}.
We strictly follow ~\cite{zhuge2021kaleido} constructing its image-text retrieval task.
In its text-to-image retrieval, the model is required to pick the matched image from 101 images given a text. 
In the 101 images, one is positively paired with the text and the other 100 are randomly paired but sharing the same sub-category as the positive, increasing the difficulty.
The same setting is for its image-to-text retrieval.
We fine-tune our smallest ResNet50-based 6-layer CommerceMM (L9 in Table~\ref{table:ablation_study}) on the dataset with contrastive learning.
In Table.~\ref{tab:fashiongen}, we compare our model with the state-of-art commerce-domain pre-trained models~\cite{gao2020fashionbert, zhuge2021kaleido}.
We found even our smallest model already outperforms~\cite{gao2020fashionbert, zhuge2021kaleido} with a clear margin, indicating CommerceMM's superior transferability.
\begin{table}[t!]
\centering
\small
\begin{tabular}{l | c c c | c c c}
& \multicolumn{3}{c|}{FashionGen T2I} & \multicolumn{3}{c}{FashionGen I2T} \\
& R1 & R5 & R10 & R1 & R5 & R10 \\
\hline 
FashionBERT~\cite{gao2020fashionbert} & 26.8 & 46.5 & 55.7 & 24.0 & 46.3 & 52.1 \\
KaleidoBERT~\cite{zhuge2021kaleido} & 33.9 & 60.6 & 68.6 & 28.0 & 60.1 & 68.4 \\
CommerceMM (small) & \bf{39.6} & \bf{61.5} & \bf{72.7} & \bf{41.6} & \bf{64.0} & \bf{72.8} \\
\hline
\end{tabular}
\caption{Image-Text Retrieval on FashionGen~\cite{rostamzadeh2018fashion}.}\label{tab:fashiongen}
\end{table}

\subsection{Modality Randomization}
\label{sec:random_k}
As in Sec.~\ref{sec:model_overview}, our model design allows us to dynamically change the text encoder and multimodal fusion encoder by assigning different $K$ and $M$ layers to each.
Previous works~\cite{wang2021vlmo, wang2021ufo, you2021ma, akbari2021vatt} show the modal-agnostic training can be beneficial to the multimodal understanding.
Our approach follows the same spirit.
At each training step, we randomly assign $K$ and $M$ transformer layers (while keeping the sum of $K$ and $M$ unchanged) to the text encoder and multimodal encoder, so that every layer can share the knowledge from both text and multimodal modalities.

\begin{figure}
    \centering
    \includegraphics[width=1.0\linewidth]{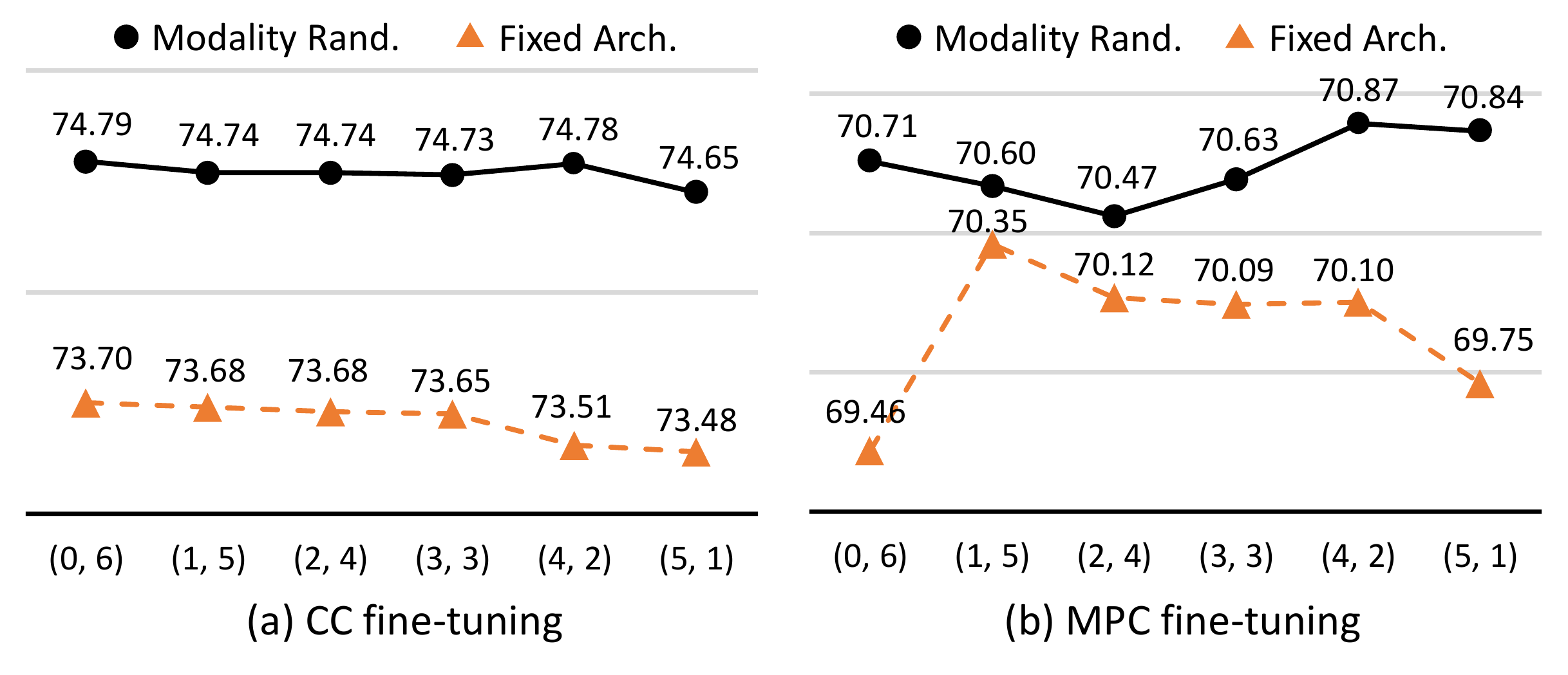}
    \vspace{-0.5cm}
    \caption{Effect of (K, M) from a fixed-architecture (3, 3) pre-trained model and modality-randomized pre-trained model. (K, M) stands for (\#text layers, \#multimodal layers).}\label{table:random_k}
\end{figure}

To validate this interesting idea, we set up a light-weighted 2nd-stage pre-training for the modality randomization with 5 image-text pre-training tasks.
We then fine-tune our model on CC and MPC, both measuring the multimodal recognition capability.
Fig.~\ref{table:random_k} compares the fixed-architecture and modality-randomized pre-trained models.
Typically the pre-training and fine-tuning models should share the same architecture, but we propose to change the model architecture at fine-tuning stage for potential better performance. 
We observe the modality-randomized pre-training brings better performance than fixed-arch under any architecture setting at fine-tuning.
We also found a smaller accuracy variance of the modality-randomized model under different architectures, showing its robustness to the model change.

\begin{figure}
    \centering
    \includegraphics[width=0.53\linewidth]{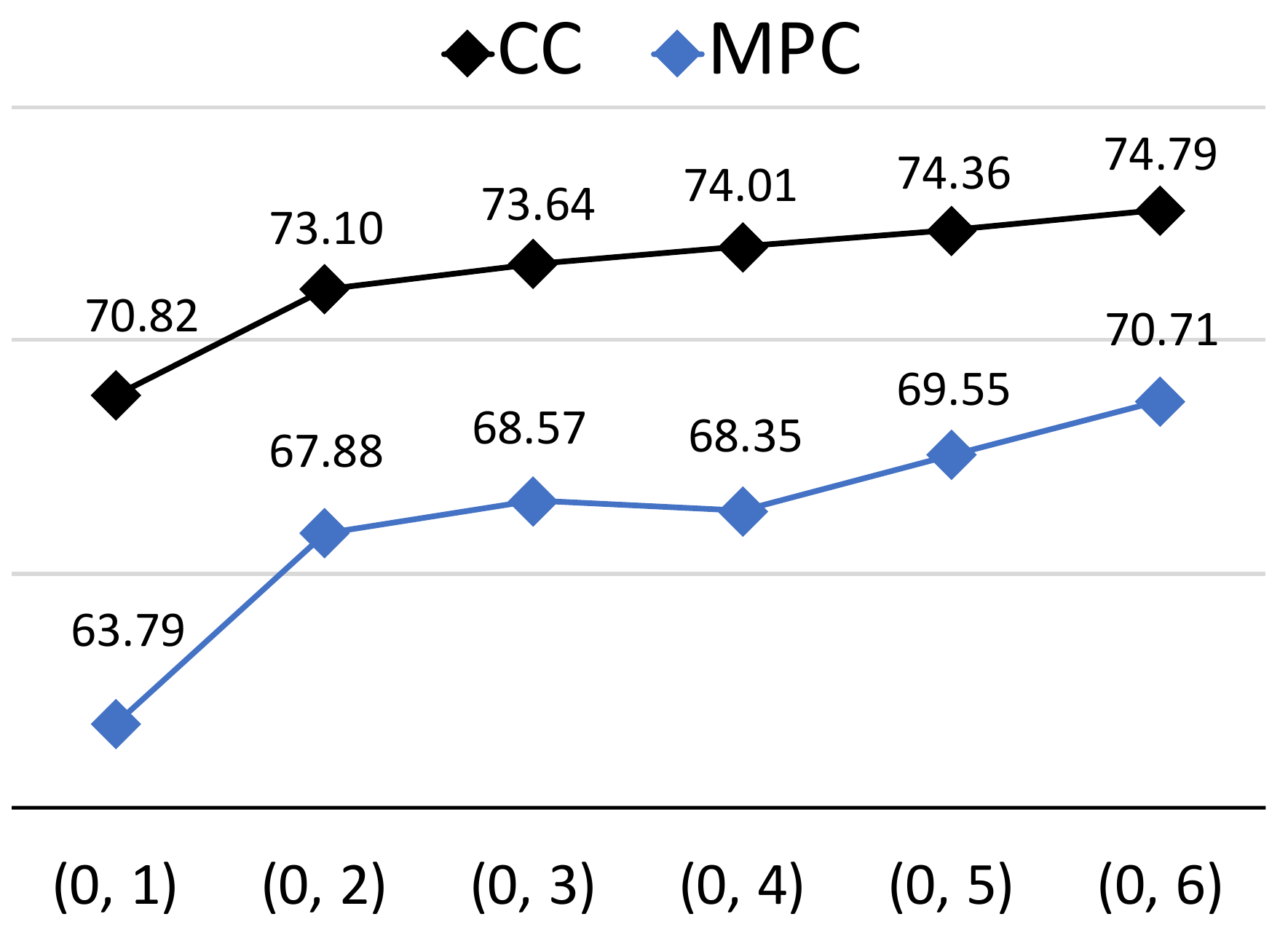}
    \vspace{-0.3cm}
    \caption{Effect of total number of layers from a modality-randomized model evaluated on CC and MPC after fine-tuning. (K, M) stands for (\#text layers, \#multimodal layers). K=0 means early-fusion model without text encoder.}\label{table:random_k_fusion_layer}
\end{figure}

We also experiment with changing the total number of layers.
In Figure~\ref{table:random_k_fusion_layer}, we plot the CC and MPC accuracy with $K=0$ text encoder and different $M$ fusion layers (each is an early fusion model without text encoder).
While it seems no surprise the deeper model brings better accuracy, we found our small models also perform well without notable performance drop compared with using full layers.
Note our 2-layer model achieves 73.10 on CC and 67.88 on MPC, which are already better than the 6-layer model without pre-training in L1 of Table~\ref{table:ablation_study}.
Thus with modality randomization, we can flexibly adjust our model architecture to satisfy different efficiency constraints in practice.

Our modality randomization can be further explored by incorporating it into the proposed pre-training in 1 stage.
We leave this promising idea to future work.

\subsection{Visualization}
\begin{figure}[h!]
    \centering
    \includegraphics[width=0.9\linewidth]{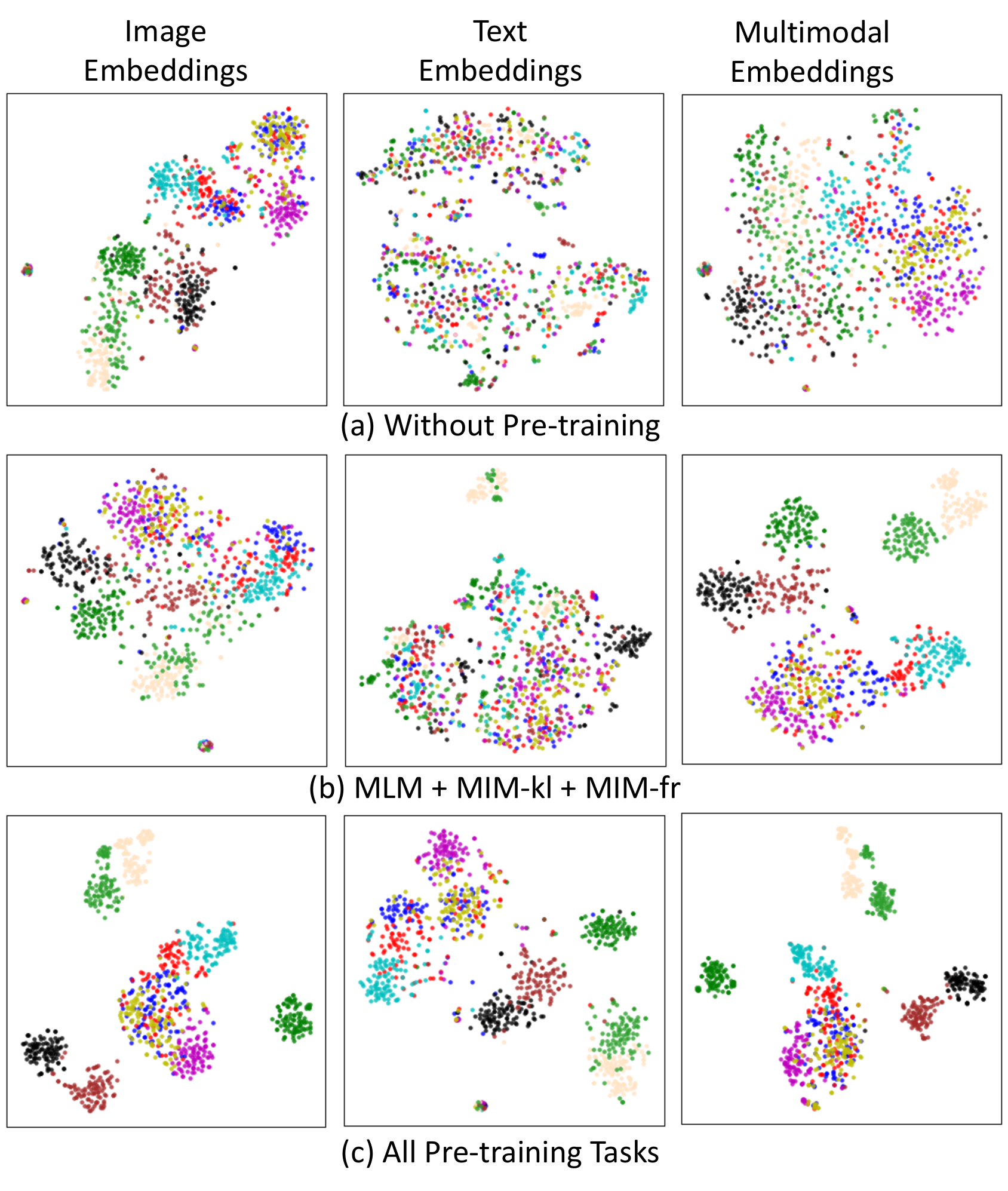}
    \vspace{-.4cm}
    \caption{T-SNE of different pre-trained models' embeddings.}
    \label{fig:embedding_visualization}
\end{figure}

We visualize the embeddings from the image encoder, text encoder, and multimodal fusion encoder respectively in Fig.~\ref{fig:embedding_visualization}.
Specifically, we feed 1K multimodal catalog posts from 10 most popular categories into CommerceMM.
T-SNE is applied for the visualization and the colors are corresponding to the annotated categories.
We compare the results of without pre-training (initialized from ResNet+XLM-R), image-text pre-training with MLM, MIM-kl, and MIM-fr, and full 14-task pre-training.
We observe that with additional retrieval tasks in pre-training, \eg, ITC and Omni, the embeddings of the same class are better clustered with closer distance.
This indicates our proposed pre-training tasks helps learning a more discriminative representation for each modality.

\begin{figure}[h!]
    \centering
    \includegraphics[width=1.0\linewidth]{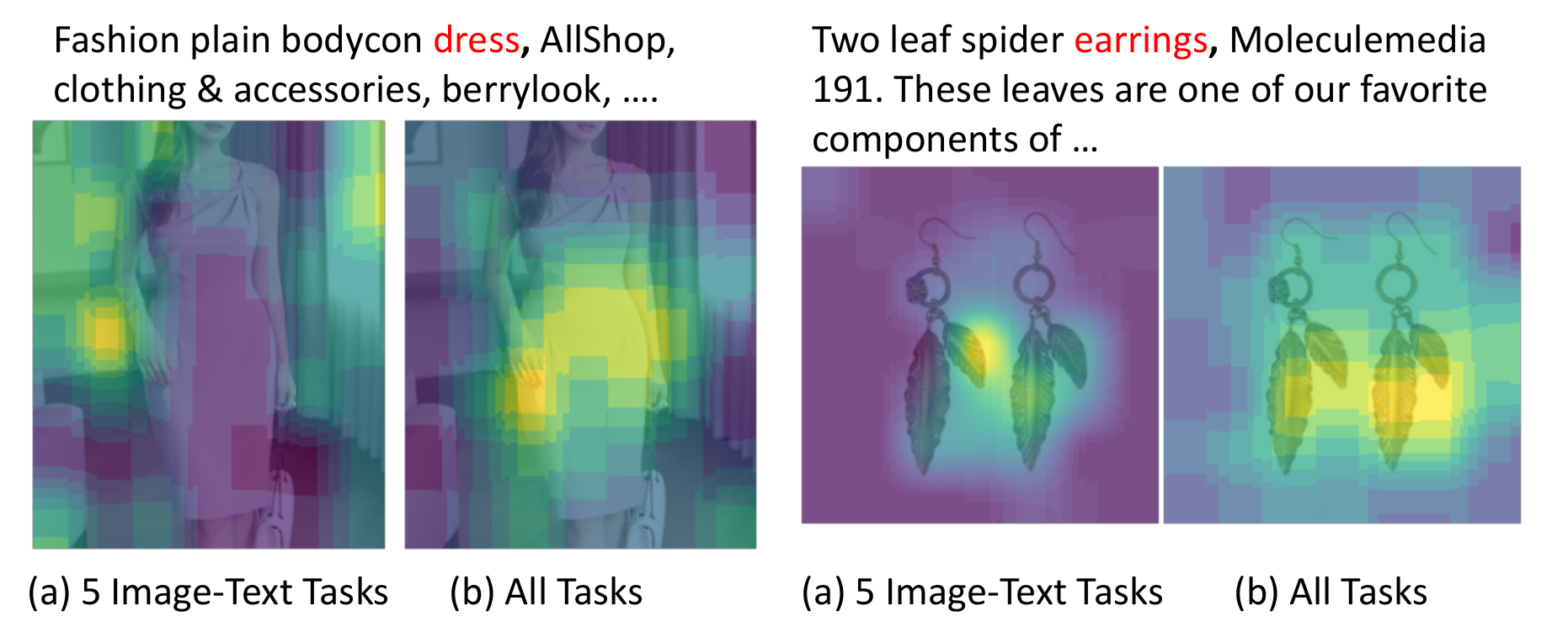}
    \caption{Visualization of text-to-image attention from different pre-trained models.}
    \label{fig:attn_visualization}
\end{figure}

We also visualize the text-to-image attention in Fig.~\ref{fig:attn_visualization}.
Comparing the the models pre-trained with 5 image-text tasks and all tasks (with Omni-Retrieval pre-training), we observe the cross-modal attention from all pre-training can better attend to the right regions referred by the key words, \eg, ``dress" and ``earrings".

\subsection{Deployment and Product Impacts}\label{sec:deployment}
An early version of CommerceMM model has been deployed at Facebook, \ie, L2 in Table~\ref{tab:model_archi}. 
We pre-trained the GrokNet~\cite{bell2020groknet}-based 6-layer model using a subset of the tasks (MLM, MIM-kl, MIM-fr and ITM) and then fine-tuned on three downstream tasks. 
We confirmed the benefits of this framework by running A/B test experiments on Catalog Categorization (CC), Marketplace Categorization (MPC) and Catalog Attributes applications.

\noindent\textbf{Catalog Category Filters} Users on Shops have access to category filters which lets them search products for a specific category within a shop. With our newly launched model, we have increased the number of shops with category filter by over 4 times. 

\noindent\textbf{Catalog Attributes Filters} 
Attributes are fine-grained characteristic of objects, such as color, pattern, material, etc. 
We ran A/B tests to confirm the improvement of attributes prediction on IG and FB Shops for three specific product attributes - color, gender and material. 
This model enabled a launch of these three attribute filters on 52.7\% of all Shops on the platform due to significant precision and coverage improvements compared to the baseline production models. 

After inference, all these category and attributes predictions are stored in a distributed key-value store, which gets consumed by multiple product groups across the company to improve various ranking and recommendation products. 

This early deployed version gives a substantial evidence that the pre-training is helpful for downstream tasks. The end-to-end CommerceMM will be deployed in the same way with a simplification of removing the service call of GrokNet Hash~\cite{bell2020groknet}, \ie, the inference will be more efficient.


\section{Conclusion}
We introduce CommerceMM, a large-scale commerce multimodal model at Facebook.
We present 5 effective image-text pre-training tasks, and propose a novel set of Omni-Retrieval tasks on cross-modal and cross-pair data.
Pre-trained on the large-scale diverse multi-lingual multimodal data, CommerceMM outperforms state-of-art models across 7 downstream tasks with a large margin.
An early version of the model has been deployed at Facebook with significant product impacts.

\begin{acks}
We thank our colleagues Wenliang Gao and Maolong Tang for the in-depth discussion and feedback; and Yuyu Zhu, Xueting Yan, Kartikay Khandelwal, and Yang Bai for supporting an early version of this pre-training in another domain.
We also thank Amanpreet Singh, Vedanuj Goswami, Sasha Sheng and Ronghang Hu for both development and deployment support of MMF.
\end{acks}

\bibliographystyle{ACM-Reference-Format}
\bibliography{bib}


\begin{thebibliography}{51}


\ifx \showCODEN    \undefined \def \showCODEN     #1{\unskip}     \fi
\ifx \showDOI      \undefined \def \showDOI       #1{#1}\fi
\ifx \showISBNx    \undefined \def \showISBNx     #1{\unskip}     \fi
\ifx \showISBNxiii \undefined \def \showISBNxiii  #1{\unskip}     \fi
\ifx \showISSN     \undefined \def \showISSN      #1{\unskip}     \fi
\ifx \showLCCN     \undefined \def \showLCCN      #1{\unskip}     \fi
\ifx \shownote     \undefined \def \shownote      #1{#1}          \fi
\ifx \showarticletitle \undefined \def \showarticletitle #1{#1}   \fi
\ifx \showURL      \undefined \def \showURL       {\relax}        \fi
\providecommand\bibfield[2]{#2}
\providecommand\bibinfo[2]{#2}
\providecommand\natexlab[1]{#1}
\providecommand\showeprint[2][]{arXiv:#2}

\bibitem[\protect\citeauthoryear{Akbari, Yuan, Qian, Chuang, Chang, Cui, and
  Gong}{Akbari et~al\mbox{.}}{2021}]%
        {akbari2021vatt}
\bibfield{author}{\bibinfo{person}{Hassan Akbari}, \bibinfo{person}{Liangzhe
  Yuan}, \bibinfo{person}{Rui Qian}, \bibinfo{person}{Wei-Hong Chuang},
  \bibinfo{person}{Shih-Fu Chang}, \bibinfo{person}{Yin Cui}, {and}
  \bibinfo{person}{Boqing Gong}.} \bibinfo{year}{2021}\natexlab{}.
\newblock \showarticletitle{Vatt: Transformers for multimodal self-supervised
  learning from raw video, audio and text}.
\newblock \bibinfo{journal}{\emph{NeurIPS}} (\bibinfo{year}{2021}).
\newblock


\bibitem[\protect\citeauthoryear{Anderson, He, Buehler, Teney, Johnson, Gould,
  and Zhang}{Anderson et~al\mbox{.}}{2018}]%
        {anderson2018bottom}
\bibfield{author}{\bibinfo{person}{Peter Anderson}, \bibinfo{person}{Xiaodong
  He}, \bibinfo{person}{Chris Buehler}, \bibinfo{person}{Damien Teney},
  \bibinfo{person}{Mark Johnson}, \bibinfo{person}{Stephen Gould}, {and}
  \bibinfo{person}{Lei Zhang}.} \bibinfo{year}{2018}\natexlab{}.
\newblock \showarticletitle{Bottom-up and top-down attention for image
  captioning and visual question answering}. In
  \bibinfo{booktitle}{\emph{CVPR}}.
\newblock


\bibitem[\protect\citeauthoryear{Antol, Agrawal, Lu, Mitchell, Batra, Zitnick,
  and Parikh}{Antol et~al\mbox{.}}{2015}]%
        {antol2015vqa}
\bibfield{author}{\bibinfo{person}{Stanislaw Antol}, \bibinfo{person}{Aishwarya
  Agrawal}, \bibinfo{person}{Jiasen Lu}, \bibinfo{person}{Margaret Mitchell},
  \bibinfo{person}{Dhruv Batra}, \bibinfo{person}{C~Lawrence Zitnick}, {and}
  \bibinfo{person}{Devi Parikh}.} \bibinfo{year}{2015}\natexlab{}.
\newblock \showarticletitle{Vqa: Visual question answering}. In
  \bibinfo{booktitle}{\emph{ICCV}}.
\newblock


\bibitem[\protect\citeauthoryear{Bell, Liu, Alsheikh, Tang, Pizzi, Henning,
  Singh, Parkhi, and Borisyuk}{Bell et~al\mbox{.}}{2020}]%
        {bell2020groknet}
\bibfield{author}{\bibinfo{person}{Sean Bell}, \bibinfo{person}{Yiqun Liu},
  \bibinfo{person}{Sami Alsheikh}, \bibinfo{person}{Yina Tang},
  \bibinfo{person}{Edward Pizzi}, \bibinfo{person}{M Henning},
  \bibinfo{person}{Karun Singh}, \bibinfo{person}{Omkar Parkhi}, {and}
  \bibinfo{person}{Fedor Borisyuk}.} \bibinfo{year}{2020}\natexlab{}.
\newblock \showarticletitle{Groknet: Unified computer vision model trunk and
  embeddings for commerce}. In \bibinfo{booktitle}{\emph{KDD}}.
\newblock


\bibitem[\protect\citeauthoryear{Chen, Li, Yu, Kholy, Ahmed, Gan, Cheng, and
  Liu}{Chen et~al\mbox{.}}{2020}]%
        {chen2019uniter}
\bibfield{author}{\bibinfo{person}{Yen-Chun Chen}, \bibinfo{person}{Linjie Li},
  \bibinfo{person}{Licheng Yu}, \bibinfo{person}{Ahmed~El Kholy},
  \bibinfo{person}{Faisal Ahmed}, \bibinfo{person}{Zhe Gan},
  \bibinfo{person}{Yu Cheng}, {and} \bibinfo{person}{Jingjing Liu}.}
  \bibinfo{year}{2020}\natexlab{}.
\newblock \showarticletitle{Uniter: Learning universal image-text
  representations}. In \bibinfo{booktitle}{\emph{ECCV}}.
\newblock


\bibitem[\protect\citeauthoryear{Conneau, Khandelwal, Goyal, Chaudhary, Wenzek,
  Guzm{\'a}n, Grave, Ott, Zettlemoyer, and Stoyanov}{Conneau
  et~al\mbox{.}}{2019}]%
        {conneau2019unsupervised}
\bibfield{author}{\bibinfo{person}{Alexis Conneau}, \bibinfo{person}{Kartikay
  Khandelwal}, \bibinfo{person}{Naman Goyal}, \bibinfo{person}{Vishrav
  Chaudhary}, \bibinfo{person}{Guillaume Wenzek}, \bibinfo{person}{Francisco
  Guzm{\'a}n}, \bibinfo{person}{Edouard Grave}, \bibinfo{person}{Myle Ott},
  \bibinfo{person}{Luke Zettlemoyer}, {and} \bibinfo{person}{Veselin
  Stoyanov}.} \bibinfo{year}{2019}\natexlab{}.
\newblock \showarticletitle{Unsupervised cross-lingual representation learning
  at scale}.
\newblock \bibinfo{journal}{\emph{arXiv preprint arXiv:1911.02116}}
  (\bibinfo{year}{2019}).
\newblock


\bibitem[\protect\citeauthoryear{Devlin, Chang, Lee, and Toutanova}{Devlin
  et~al\mbox{.}}{2018}]%
        {devlin2018bert}
\bibfield{author}{\bibinfo{person}{Jacob Devlin}, \bibinfo{person}{Ming-Wei
  Chang}, \bibinfo{person}{Kenton Lee}, {and} \bibinfo{person}{Kristina
  Toutanova}.} \bibinfo{year}{2018}\natexlab{}.
\newblock \showarticletitle{Bert: Pre-training of deep bidirectional
  transformers for language understanding}. In
  \bibinfo{booktitle}{\emph{NAACL}}.
\newblock


\bibitem[\protect\citeauthoryear{Dong, Zhan, Wu, Wei, Wei, Lu, and Liang}{Dong
  et~al\mbox{.}}{2021}]%
        {dong2021m5product}
\bibfield{author}{\bibinfo{person}{Xiao Dong}, \bibinfo{person}{Xunlin Zhan},
  \bibinfo{person}{Yangxin Wu}, \bibinfo{person}{Yunchao Wei},
  \bibinfo{person}{Xiaoyong Wei}, \bibinfo{person}{Minlong Lu}, {and}
  \bibinfo{person}{Xiaodan Liang}.} \bibinfo{year}{2021}\natexlab{}.
\newblock \showarticletitle{M5product: A multi-modal pretraining benchmark for
  e-commercial product downstream tasks}.
\newblock \bibinfo{journal}{\emph{arXiv preprint arXiv:2109.04275}}
  (\bibinfo{year}{2021}).
\newblock


\bibitem[\protect\citeauthoryear{Dosovitskiy, Beyer, Kolesnikov, Weissenborn,
  Zhai, Unterthiner, Dehghani, Minderer, Heigold, Gelly,
  et~al\mbox{.}}{Dosovitskiy et~al\mbox{.}}{2020}]%
        {dosovitskiy2020image}
\bibfield{author}{\bibinfo{person}{Alexey Dosovitskiy}, \bibinfo{person}{Lucas
  Beyer}, \bibinfo{person}{Alexander Kolesnikov}, \bibinfo{person}{Dirk
  Weissenborn}, \bibinfo{person}{Xiaohua Zhai}, \bibinfo{person}{Thomas
  Unterthiner}, \bibinfo{person}{Mostafa Dehghani}, \bibinfo{person}{Matthias
  Minderer}, \bibinfo{person}{Georg Heigold}, \bibinfo{person}{Sylvain Gelly},
  {et~al\mbox{.}}} \bibinfo{year}{2020}\natexlab{}.
\newblock \showarticletitle{An image is worth 16x16 words: Transformers for
  image recognition at scale}.
\newblock \bibinfo{journal}{\emph{arXiv preprint arXiv:2010.11929}}
  (\bibinfo{year}{2020}).
\newblock


\bibitem[\protect\citeauthoryear{Dou, Xu, Gan, Wang, Wang, Wang, Zhu, Liu,
  Zeng, et~al\mbox{.}}{Dou et~al\mbox{.}}{2021}]%
        {dou2021empirical}
\bibfield{author}{\bibinfo{person}{Zi-Yi Dou}, \bibinfo{person}{Yichong Xu},
  \bibinfo{person}{Zhe Gan}, \bibinfo{person}{Jianfeng Wang},
  \bibinfo{person}{Shuohang Wang}, \bibinfo{person}{Lijuan Wang},
  \bibinfo{person}{Chenguang Zhu}, \bibinfo{person}{Zicheng Liu},
  \bibinfo{person}{Michael Zeng}, {et~al\mbox{.}}}
  \bibinfo{year}{2021}\natexlab{}.
\newblock \showarticletitle{An Empirical Study of Training End-to-End
  Vision-and-Language Transformers}.
\newblock \bibinfo{journal}{\emph{arXiv preprint arXiv:2111.02387}}
  (\bibinfo{year}{2021}).
\newblock


\bibitem[\protect\citeauthoryear{Gao, Jin, Chen, Qiu, Li, Wei, Hu, and
  Wang}{Gao et~al\mbox{.}}{2020}]%
        {gao2020fashionbert}
\bibfield{author}{\bibinfo{person}{Dehong Gao}, \bibinfo{person}{Linbo Jin},
  \bibinfo{person}{Ben Chen}, \bibinfo{person}{Minghui Qiu},
  \bibinfo{person}{Peng Li}, \bibinfo{person}{Yi Wei}, \bibinfo{person}{Yi Hu},
  {and} \bibinfo{person}{Hao Wang}.} \bibinfo{year}{2020}\natexlab{}.
\newblock \showarticletitle{Fashionbert: Text and image matching with adaptive
  loss for cross-modal retrieval}. In \bibinfo{booktitle}{\emph{SIGIR}}.
\newblock


\bibitem[\protect\citeauthoryear{He, Chen, Xie, Li, Doll{\'a}r, and
  Girshick}{He et~al\mbox{.}}{2021}]%
        {he2021masked}
\bibfield{author}{\bibinfo{person}{Kaiming He}, \bibinfo{person}{Xinlei Chen},
  \bibinfo{person}{Saining Xie}, \bibinfo{person}{Yanghao Li},
  \bibinfo{person}{Piotr Doll{\'a}r}, {and} \bibinfo{person}{Ross Girshick}.}
  \bibinfo{year}{2021}\natexlab{}.
\newblock \showarticletitle{Masked autoencoders are scalable vision learners}.
\newblock \bibinfo{journal}{\emph{arXiv preprint arXiv:2111.06377}}
  (\bibinfo{year}{2021}).
\newblock


\bibitem[\protect\citeauthoryear{Hu, Yin, Lin, Wang, Zhang, Gao, and Liu}{Hu
  et~al\mbox{.}}{2021}]%
        {hu2020vivo}
\bibfield{author}{\bibinfo{person}{Xiaowei Hu}, \bibinfo{person}{Xi Yin},
  \bibinfo{person}{Kevin Lin}, \bibinfo{person}{Lijuan Wang},
  \bibinfo{person}{Lei Zhang}, \bibinfo{person}{Jianfeng Gao}, {and}
  \bibinfo{person}{Zicheng Liu}.} \bibinfo{year}{2021}\natexlab{}.
\newblock \showarticletitle{Vivo: Surpassing human performance in novel object
  captioning with visual vocabulary pre-training}. In
  \bibinfo{booktitle}{\emph{AAAI}}.
\newblock


\bibitem[\protect\citeauthoryear{Huang, Zeng, Liu, Fu, and Fu}{Huang
  et~al\mbox{.}}{2020}]%
        {huang2020pixel}
\bibfield{author}{\bibinfo{person}{Zhicheng Huang}, \bibinfo{person}{Zhaoyang
  Zeng}, \bibinfo{person}{Bei Liu}, \bibinfo{person}{Dongmei Fu}, {and}
  \bibinfo{person}{Jianlong Fu}.} \bibinfo{year}{2020}\natexlab{}.
\newblock \showarticletitle{Pixel-bert: Aligning image pixels with text by deep
  multi-modal transformers}.
\newblock \bibinfo{journal}{\emph{arXiv preprint arXiv:2004.00849}}
  (\bibinfo{year}{2020}).
\newblock


\bibitem[\protect\citeauthoryear{Jia, Yang, Xia, Chen, Parekh, Pham, Le, Sung,
  Li, and Duerig}{Jia et~al\mbox{.}}{2021}]%
        {jia2021scaling}
\bibfield{author}{\bibinfo{person}{Chao Jia}, \bibinfo{person}{Yinfei Yang},
  \bibinfo{person}{Ye Xia}, \bibinfo{person}{Yi-Ting Chen},
  \bibinfo{person}{Zarana Parekh}, \bibinfo{person}{Hieu Pham},
  \bibinfo{person}{Quoc Le}, \bibinfo{person}{Yun-Hsuan Sung},
  \bibinfo{person}{Zhen Li}, {and} \bibinfo{person}{Tom Duerig}.}
  \bibinfo{year}{2021}\natexlab{}.
\newblock \showarticletitle{Scaling up visual and vision-language
  representation learning with noisy text supervision}. In
  \bibinfo{booktitle}{\emph{ICML}}.
\newblock


\bibitem[\protect\citeauthoryear{Kim, Son, and Kim}{Kim et~al\mbox{.}}{2021}]%
        {kim2021vilt}
\bibfield{author}{\bibinfo{person}{Wonjae Kim}, \bibinfo{person}{Bokyung Son},
  {and} \bibinfo{person}{Ildoo Kim}.} \bibinfo{year}{2021}\natexlab{}.
\newblock \showarticletitle{Vilt: Vision-and-language transformer without
  convolution or region supervision}. In \bibinfo{booktitle}{\emph{ICML}}.
\newblock


\bibitem[\protect\citeauthoryear{Krishna, Zhu, Groth, Johnson, Hata, Kravitz,
  Chen, Kalantidis, Li, Shamma, et~al\mbox{.}}{Krishna et~al\mbox{.}}{2017}]%
        {krishna2017visual}
\bibfield{author}{\bibinfo{person}{Ranjay Krishna}, \bibinfo{person}{Yuke Zhu},
  \bibinfo{person}{Oliver Groth}, \bibinfo{person}{Justin Johnson},
  \bibinfo{person}{Kenji Hata}, \bibinfo{person}{Joshua Kravitz},
  \bibinfo{person}{Stephanie Chen}, \bibinfo{person}{Yannis Kalantidis},
  \bibinfo{person}{Li-Jia Li}, \bibinfo{person}{David~A Shamma},
  {et~al\mbox{.}}} \bibinfo{year}{2017}\natexlab{}.
\newblock \showarticletitle{Visual genome: Connecting language and vision using
  crowdsourced dense image annotations}.
\newblock \bibinfo{journal}{\emph{IJCV}} (\bibinfo{year}{2017}).
\newblock


\bibitem[\protect\citeauthoryear{Li, Duan, Fang, Gong, and Jiang}{Li
  et~al\mbox{.}}{2020b}]%
        {li2020unicoder}
\bibfield{author}{\bibinfo{person}{Gen Li}, \bibinfo{person}{Nan Duan},
  \bibinfo{person}{Yuejian Fang}, \bibinfo{person}{Ming Gong}, {and}
  \bibinfo{person}{Daxin Jiang}.} \bibinfo{year}{2020}\natexlab{b}.
\newblock \showarticletitle{Unicoder-vl: A universal encoder for vision and
  language by cross-modal pre-training}. In \bibinfo{booktitle}{\emph{AAAI}}.
\newblock


\bibitem[\protect\citeauthoryear{Li, Selvaraju, Gotmare, Joty, Xiong, and
  Hoi}{Li et~al\mbox{.}}{2021}]%
        {li2021align}
\bibfield{author}{\bibinfo{person}{Junnan Li}, \bibinfo{person}{Ramprasaath
  Selvaraju}, \bibinfo{person}{Akhilesh Gotmare}, \bibinfo{person}{Shafiq
  Joty}, \bibinfo{person}{Caiming Xiong}, {and} \bibinfo{person}{Steven
  Chu~Hong Hoi}.} \bibinfo{year}{2021}\natexlab{}.
\newblock \showarticletitle{Align before fuse: Vision and language
  representation learning with momentum distillation}.
\newblock \bibinfo{journal}{\emph{NeurIPS}} (\bibinfo{year}{2021}).
\newblock


\bibitem[\protect\citeauthoryear{Li, Chen, Cheng, Gan, Yu, and Liu}{Li
  et~al\mbox{.}}{2020a}]%
        {li2020hero}
\bibfield{author}{\bibinfo{person}{Linjie Li}, \bibinfo{person}{Yen-Chun Chen},
  \bibinfo{person}{Yu Cheng}, \bibinfo{person}{Zhe Gan},
  \bibinfo{person}{Licheng Yu}, {and} \bibinfo{person}{Jingjing Liu}.}
  \bibinfo{year}{2020}\natexlab{a}.
\newblock \showarticletitle{Hero: Hierarchical encoder for video+ language
  omni-representation pre-training}. In \bibinfo{booktitle}{\emph{EMNLP}}.
\newblock


\bibitem[\protect\citeauthoryear{Li, Yatskar, Yin, Hsieh, and Chang}{Li
  et~al\mbox{.}}{2019}]%
        {li2019visualbert}
\bibfield{author}{\bibinfo{person}{Liunian~Harold Li}, \bibinfo{person}{Mark
  Yatskar}, \bibinfo{person}{Da Yin}, \bibinfo{person}{Cho-Jui Hsieh}, {and}
  \bibinfo{person}{Kai-Wei Chang}.} \bibinfo{year}{2019}\natexlab{}.
\newblock \showarticletitle{Visualbert: A simple and performant baseline for
  vision and language}.
\newblock \bibinfo{journal}{\emph{arXiv preprint arXiv:1908.03557}}
  (\bibinfo{year}{2019}).
\newblock


\bibitem[\protect\citeauthoryear{Li, Gao, Niu, Xiao, Liu, Liu, Wu, and Wang}{Li
  et~al\mbox{.}}{2020c}]%
        {li2020unimo}
\bibfield{author}{\bibinfo{person}{Wei Li}, \bibinfo{person}{Can Gao},
  \bibinfo{person}{Guocheng Niu}, \bibinfo{person}{Xinyan Xiao},
  \bibinfo{person}{Hao Liu}, \bibinfo{person}{Jiachen Liu},
  \bibinfo{person}{Hua Wu}, {and} \bibinfo{person}{Haifeng Wang}.}
  \bibinfo{year}{2020}\natexlab{c}.
\newblock \showarticletitle{Unimo: Towards unified-modal understanding and
  generation via cross-modal contrastive learning}.
\newblock \bibinfo{journal}{\emph{arXiv preprint arXiv:2012.15409}}
  (\bibinfo{year}{2020}).
\newblock


\bibitem[\protect\citeauthoryear{Li, Yin, Li, Zhang, Hu, Zhang, Wang, Hu, Dong,
  Wei, et~al\mbox{.}}{Li et~al\mbox{.}}{2020d}]%
        {li2020oscar}
\bibfield{author}{\bibinfo{person}{Xiujun Li}, \bibinfo{person}{Xi Yin},
  \bibinfo{person}{Chunyuan Li}, \bibinfo{person}{Pengchuan Zhang},
  \bibinfo{person}{Xiaowei Hu}, \bibinfo{person}{Lei Zhang},
  \bibinfo{person}{Lijuan Wang}, \bibinfo{person}{Houdong Hu},
  \bibinfo{person}{Li Dong}, \bibinfo{person}{Furu Wei}, {et~al\mbox{.}}}
  \bibinfo{year}{2020}\natexlab{d}.
\newblock \showarticletitle{Oscar: Object-semantics aligned pre-training for
  vision-language tasks}. In \bibinfo{booktitle}{\emph{ECCV}}.
\newblock


\bibitem[\protect\citeauthoryear{Lin, Maire, Belongie, Hays, Perona, Ramanan,
  Doll{\'a}r, and Zitnick}{Lin et~al\mbox{.}}{2014}]%
        {lin2014microsoft}
\bibfield{author}{\bibinfo{person}{Tsung-Yi Lin}, \bibinfo{person}{Michael
  Maire}, \bibinfo{person}{Serge Belongie}, \bibinfo{person}{James Hays},
  \bibinfo{person}{Pietro Perona}, \bibinfo{person}{Deva Ramanan},
  \bibinfo{person}{Piotr Doll{\'a}r}, {and} \bibinfo{person}{C~Lawrence
  Zitnick}.} \bibinfo{year}{2014}\natexlab{}.
\newblock \showarticletitle{Microsoft coco: Common objects in context}. In
  \bibinfo{booktitle}{\emph{ECCV}}.
\newblock


\bibitem[\protect\citeauthoryear{Lu, Batra, Parikh, and Lee}{Lu
  et~al\mbox{.}}{2019}]%
        {lu2019vilbert}
\bibfield{author}{\bibinfo{person}{Jiasen Lu}, \bibinfo{person}{Dhruv Batra},
  \bibinfo{person}{Devi Parikh}, {and} \bibinfo{person}{Stefan Lee}.}
  \bibinfo{year}{2019}\natexlab{}.
\newblock \showarticletitle{Vilbert: Pretraining task-agnostic visiolinguistic
  representations for vision-and-language tasks}.
\newblock \bibinfo{journal}{\emph{NeurIPS}} (\bibinfo{year}{2019}).
\newblock


\bibitem[\protect\citeauthoryear{Mahajan, Girshick, Ramanathan, He, Paluri, Li,
  Bharambe, and Van Der~Maaten}{Mahajan et~al\mbox{.}}{2018}]%
        {mahajan2018exploring}
\bibfield{author}{\bibinfo{person}{Dhruv Mahajan}, \bibinfo{person}{Ross
  Girshick}, \bibinfo{person}{Vignesh Ramanathan}, \bibinfo{person}{Kaiming
  He}, \bibinfo{person}{Manohar Paluri}, \bibinfo{person}{Yixuan Li},
  \bibinfo{person}{Ashwin Bharambe}, {and} \bibinfo{person}{Laurens Van
  Der~Maaten}.} \bibinfo{year}{2018}\natexlab{}.
\newblock \showarticletitle{Exploring the limits of weakly supervised
  pretraining}. In \bibinfo{booktitle}{\emph{ECCV}}.
\newblock


\bibitem[\protect\citeauthoryear{Miech, Alayrac, Laptev, Sivic, and
  Zisserman}{Miech et~al\mbox{.}}{2021}]%
        {miech2021thinking}
\bibfield{author}{\bibinfo{person}{Antoine Miech},
  \bibinfo{person}{Jean-Baptiste Alayrac}, \bibinfo{person}{Ivan Laptev},
  \bibinfo{person}{Josef Sivic}, {and} \bibinfo{person}{Andrew Zisserman}.}
  \bibinfo{year}{2021}\natexlab{}.
\newblock \showarticletitle{Thinking Fast and Slow: Efficient Text-to-Visual
  Retrieval with Transformers}. In \bibinfo{booktitle}{\emph{CVPR}}.
\newblock


\bibitem[\protect\citeauthoryear{Ordonez, Kulkarni, and Berg}{Ordonez
  et~al\mbox{.}}{2011}]%
        {ordonez2011im2text}
\bibfield{author}{\bibinfo{person}{Vicente Ordonez}, \bibinfo{person}{Girish
  Kulkarni}, {and} \bibinfo{person}{Tamara Berg}.}
  \bibinfo{year}{2011}\natexlab{}.
\newblock \showarticletitle{Im2text: Describing images using 1 million
  captioned photographs}.
\newblock \bibinfo{journal}{\emph{NeurIPS}} (\bibinfo{year}{2011}).
\newblock


\bibitem[\protect\citeauthoryear{Plummer, Wang, Cervantes, Caicedo,
  Hockenmaier, and Lazebnik}{Plummer et~al\mbox{.}}{2015}]%
        {plummer2015flickr30k}
\bibfield{author}{\bibinfo{person}{Bryan~A Plummer}, \bibinfo{person}{Liwei
  Wang}, \bibinfo{person}{Chris~M Cervantes}, \bibinfo{person}{Juan~C Caicedo},
  \bibinfo{person}{Julia Hockenmaier}, {and} \bibinfo{person}{Svetlana
  Lazebnik}.} \bibinfo{year}{2015}\natexlab{}.
\newblock \showarticletitle{Flickr30k entities: Collecting region-to-phrase
  correspondences for richer image-to-sentence models}. In
  \bibinfo{booktitle}{\emph{ICCV}}.
\newblock


\bibitem[\protect\citeauthoryear{Radford, Kim, Hallacy, Ramesh, Goh, Agarwal,
  Sastry, Askell, Mishkin, Clark, et~al\mbox{.}}{Radford et~al\mbox{.}}{2021}]%
        {radford2021learning}
\bibfield{author}{\bibinfo{person}{Alec Radford}, \bibinfo{person}{Jong~Wook
  Kim}, \bibinfo{person}{Chris Hallacy}, \bibinfo{person}{Aditya Ramesh},
  \bibinfo{person}{Gabriel Goh}, \bibinfo{person}{Sandhini Agarwal},
  \bibinfo{person}{Girish Sastry}, \bibinfo{person}{Amanda Askell},
  \bibinfo{person}{Pamela Mishkin}, \bibinfo{person}{Jack Clark},
  {et~al\mbox{.}}} \bibinfo{year}{2021}\natexlab{}.
\newblock \showarticletitle{Learning transferable visual models from natural
  language supervision}. In \bibinfo{booktitle}{\emph{ICML}}.
\newblock


\bibitem[\protect\citeauthoryear{Rostamzadeh, Hosseini, Boquet, Stokowiec,
  Zhang, Jauvin, and Pal}{Rostamzadeh et~al\mbox{.}}{2018}]%
        {rostamzadeh2018fashion}
\bibfield{author}{\bibinfo{person}{Negar Rostamzadeh},
  \bibinfo{person}{Seyedarian Hosseini}, \bibinfo{person}{Thomas Boquet},
  \bibinfo{person}{Wojciech Stokowiec}, \bibinfo{person}{Ying Zhang},
  \bibinfo{person}{Christian Jauvin}, {and} \bibinfo{person}{Chris Pal}.}
  \bibinfo{year}{2018}\natexlab{}.
\newblock \showarticletitle{Fashion-gen: The generative fashion dataset and
  challenge}.
\newblock \bibinfo{journal}{\emph{arXiv preprint arXiv:1806.08317}}
  (\bibinfo{year}{2018}).
\newblock


\bibitem[\protect\citeauthoryear{Sharma, Ding, Goodman, and Soricut}{Sharma
  et~al\mbox{.}}{2018}]%
        {sharma2018conceptual}
\bibfield{author}{\bibinfo{person}{Piyush Sharma}, \bibinfo{person}{Nan Ding},
  \bibinfo{person}{Sebastian Goodman}, {and} \bibinfo{person}{Radu Soricut}.}
  \bibinfo{year}{2018}\natexlab{}.
\newblock \showarticletitle{Conceptual captions: A cleaned, hypernymed, image
  alt-text dataset for automatic image captioning}. In
  \bibinfo{booktitle}{\emph{ACL}}.
\newblock


\bibitem[\protect\citeauthoryear{Singh, Goswami, Natarajan, Jiang, Chen, Shah,
  Rohrbach, Batra, and Parikh}{Singh et~al\mbox{.}}{2020}]%
        {singh2020mmf}
\bibfield{author}{\bibinfo{person}{Amanpreet Singh}, \bibinfo{person}{Vedanuj
  Goswami}, \bibinfo{person}{Vivek Natarajan}, \bibinfo{person}{Yu Jiang},
  \bibinfo{person}{Xinlei Chen}, \bibinfo{person}{Meet Shah},
  \bibinfo{person}{Marcus Rohrbach}, \bibinfo{person}{Dhruv Batra}, {and}
  \bibinfo{person}{Devi Parikh}.} \bibinfo{year}{2020}\natexlab{}.
\newblock \bibinfo{title}{MMF: A multimodal framework for vision and language
  research}.
\newblock
  \bibinfo{howpublished}{\url{https://github.com/facebookresearch/mmf}}.
\newblock


\bibitem[\protect\citeauthoryear{Su, Zhu, Cao, Li, Lu, Wei, and Dai}{Su
  et~al\mbox{.}}{2019}]%
        {su2019vl}
\bibfield{author}{\bibinfo{person}{Weijie Su}, \bibinfo{person}{Xizhou Zhu},
  \bibinfo{person}{Yue Cao}, \bibinfo{person}{Bin Li}, \bibinfo{person}{Lewei
  Lu}, \bibinfo{person}{Furu Wei}, {and} \bibinfo{person}{Jifeng Dai}.}
  \bibinfo{year}{2019}\natexlab{}.
\newblock \showarticletitle{Vl-bert: Pre-training of generic visual-linguistic
  representations}.
\newblock \bibinfo{journal}{\emph{arXiv preprint arXiv:1908.08530}}
  (\bibinfo{year}{2019}).
\newblock


\bibitem[\protect\citeauthoryear{Sun, Myers, Vondrick, Murphy, and Schmid}{Sun
  et~al\mbox{.}}{2019}]%
        {sun2019videobert}
\bibfield{author}{\bibinfo{person}{Chen Sun}, \bibinfo{person}{Austin Myers},
  \bibinfo{person}{Carl Vondrick}, \bibinfo{person}{Kevin Murphy}, {and}
  \bibinfo{person}{Cordelia Schmid}.} \bibinfo{year}{2019}\natexlab{}.
\newblock \showarticletitle{Videobert: A joint model for video and language
  representation learning}. In \bibinfo{booktitle}{\emph{ICCV}}.
\newblock


\bibitem[\protect\citeauthoryear{Tan and Bansal}{Tan and Bansal}{2019}]%
        {tan2019lxmert}
\bibfield{author}{\bibinfo{person}{Hao Tan} {and} \bibinfo{person}{Mohit
  Bansal}.} \bibinfo{year}{2019}\natexlab{}.
\newblock \showarticletitle{Lxmert: Learning cross-modality encoder
  representations from transformers}. In \bibinfo{booktitle}{\emph{EMNLP}}.
\newblock


\bibitem[\protect\citeauthoryear{Vaswani, Shazeer, Parmar, Uszkoreit, Jones,
  Gomez, Kaiser, and Polosukhin}{Vaswani et~al\mbox{.}}{2017}]%
        {vaswani2017attention}
\bibfield{author}{\bibinfo{person}{Ashish Vaswani}, \bibinfo{person}{Noam
  Shazeer}, \bibinfo{person}{Niki Parmar}, \bibinfo{person}{Jakob Uszkoreit},
  \bibinfo{person}{Llion Jones}, \bibinfo{person}{Aidan~N Gomez},
  \bibinfo{person}{{\L}ukasz Kaiser}, {and} \bibinfo{person}{Illia
  Polosukhin}.} \bibinfo{year}{2017}\natexlab{}.
\newblock \showarticletitle{Attention is all you need}.
\newblock \bibinfo{journal}{\emph{NeurIPS}} (\bibinfo{year}{2017}).
\newblock


\bibitem[\protect\citeauthoryear{Wang, Hu, Gan, Yang, Dai, Liu, Lu, and
  Wang}{Wang et~al\mbox{.}}{2021b}]%
        {wang2021ufo}
\bibfield{author}{\bibinfo{person}{Jianfeng Wang}, \bibinfo{person}{Xiaowei
  Hu}, \bibinfo{person}{Zhe Gan}, \bibinfo{person}{Zhengyuan Yang},
  \bibinfo{person}{Xiyang Dai}, \bibinfo{person}{Zicheng Liu},
  \bibinfo{person}{Yumao Lu}, {and} \bibinfo{person}{Lijuan Wang}.}
  \bibinfo{year}{2021}\natexlab{b}.
\newblock \showarticletitle{UFO: A UniFied TransfOrmer for Vision-Language
  Representation Learning}.
\newblock \bibinfo{journal}{\emph{arXiv preprint arXiv:2111.10023}}
  (\bibinfo{year}{2021}).
\newblock


\bibitem[\protect\citeauthoryear{Wang, Bao, Dong, and Wei}{Wang
  et~al\mbox{.}}{2021a}]%
        {wang2021vlmo}
\bibfield{author}{\bibinfo{person}{Wenhui Wang}, \bibinfo{person}{Hangbo Bao},
  \bibinfo{person}{Li Dong}, {and} \bibinfo{person}{Furu Wei}.}
  \bibinfo{year}{2021}\natexlab{a}.
\newblock \showarticletitle{VLMo: Unified Vision-Language Pre-Training with
  Mixture-of-Modality-Experts}.
\newblock \bibinfo{journal}{\emph{arXiv preprint arXiv:2111.02358}}
  (\bibinfo{year}{2021}).
\newblock


\bibitem[\protect\citeauthoryear{Wang, Yu, Yu, Dai, Tsvetkov, and Cao}{Wang
  et~al\mbox{.}}{2021c}]%
        {wang2021simvlm}
\bibfield{author}{\bibinfo{person}{Zirui Wang}, \bibinfo{person}{Jiahui Yu},
  \bibinfo{person}{Adams~Wei Yu}, \bibinfo{person}{Zihang Dai},
  \bibinfo{person}{Yulia Tsvetkov}, {and} \bibinfo{person}{Yuan Cao}.}
  \bibinfo{year}{2021}\natexlab{c}.
\newblock \showarticletitle{Simvlm: Simple visual language model pretraining
  with weak supervision}.
\newblock \bibinfo{journal}{\emph{arXiv preprint arXiv:2108.10904}}
  (\bibinfo{year}{2021}).
\newblock


\bibitem[\protect\citeauthoryear{Wei, Fan, Xie, Wu, Yuille, and
  Feichtenhofer}{Wei et~al\mbox{.}}{2021}]%
        {wei2021masked}
\bibfield{author}{\bibinfo{person}{Chen Wei}, \bibinfo{person}{Haoqi Fan},
  \bibinfo{person}{Saining Xie}, \bibinfo{person}{Chao-Yuan Wu},
  \bibinfo{person}{Alan Yuille}, {and} \bibinfo{person}{Christoph
  Feichtenhofer}.} \bibinfo{year}{2021}\natexlab{}.
\newblock \showarticletitle{Masked Feature Prediction for Self-Supervised
  Visual Pre-Training}.
\newblock \bibinfo{journal}{\emph{arXiv preprint arXiv:2112.09133}}
  (\bibinfo{year}{2021}).
\newblock


\bibitem[\protect\citeauthoryear{Xie, Zhang, Cao, Lin, Bao, Yao, Dai, and
  Hu}{Xie et~al\mbox{.}}{2021}]%
        {xie2021simmim}
\bibfield{author}{\bibinfo{person}{Zhenda Xie}, \bibinfo{person}{Zheng Zhang},
  \bibinfo{person}{Yue Cao}, \bibinfo{person}{Yutong Lin},
  \bibinfo{person}{Jianmin Bao}, \bibinfo{person}{Zhuliang Yao},
  \bibinfo{person}{Qi Dai}, {and} \bibinfo{person}{Han Hu}.}
  \bibinfo{year}{2021}\natexlab{}.
\newblock \showarticletitle{Simmim: A simple framework for masked image
  modeling}.
\newblock \bibinfo{journal}{\emph{arXiv preprint arXiv:2111.09886}}
  (\bibinfo{year}{2021}).
\newblock


\bibitem[\protect\citeauthoryear{Yan, Misra, Gupta, Ghadiyaram, and
  Mahajan}{Yan et~al\mbox{.}}{2020}]%
        {yan2020clusterfit}
\bibfield{author}{\bibinfo{person}{Xueting Yan}, \bibinfo{person}{Ishan Misra},
  \bibinfo{person}{Abhinav Gupta}, \bibinfo{person}{Deepti Ghadiyaram}, {and}
  \bibinfo{person}{Dhruv Mahajan}.} \bibinfo{year}{2020}\natexlab{}.
\newblock \showarticletitle{Clusterfit: Improving generalization of visual
  representations}. In \bibinfo{booktitle}{\emph{CVPR}}.
\newblock


\bibitem[\protect\citeauthoryear{You, Zhou, Xiao, Codella, Cheng, Xu, Chang,
  and Yuan}{You et~al\mbox{.}}{2021}]%
        {you2021ma}
\bibfield{author}{\bibinfo{person}{Haoxuan You}, \bibinfo{person}{Luowei Zhou},
  \bibinfo{person}{Bin Xiao}, \bibinfo{person}{Noel~C Codella},
  \bibinfo{person}{Yu Cheng}, \bibinfo{person}{Ruochen Xu},
  \bibinfo{person}{Shih-Fu Chang}, {and} \bibinfo{person}{Lu Yuan}.}
  \bibinfo{year}{2021}\natexlab{}.
\newblock \showarticletitle{MA-CLIP: Towards Modality-Agnostic Contrastive
  Language-Image Pre-training}.
\newblock  (\bibinfo{year}{2021}).
\newblock


\bibitem[\protect\citeauthoryear{Yu, Tang, Yin, Sun, Tian, Wu, and Wang}{Yu
  et~al\mbox{.}}{2020}]%
        {yu2020ernie}
\bibfield{author}{\bibinfo{person}{Fei Yu}, \bibinfo{person}{Jiji Tang},
  \bibinfo{person}{Weichong Yin}, \bibinfo{person}{Yu Sun},
  \bibinfo{person}{Hao Tian}, \bibinfo{person}{Hua Wu}, {and}
  \bibinfo{person}{Haifeng Wang}.} \bibinfo{year}{2020}\natexlab{}.
\newblock \showarticletitle{Ernie-vil: Knowledge enhanced vision-language
  representations through scene graph}.
\newblock \bibinfo{journal}{\emph{arXiv preprint arXiv:2006.16934}}
  (\bibinfo{year}{2020}).
\newblock


\bibitem[\protect\citeauthoryear{Yu, Poirson, Yang, Berg, and Berg}{Yu
  et~al\mbox{.}}{2016}]%
        {yu2016modeling}
\bibfield{author}{\bibinfo{person}{Licheng Yu}, \bibinfo{person}{Patrick
  Poirson}, \bibinfo{person}{Shan Yang}, \bibinfo{person}{Alexander~C Berg},
  {and} \bibinfo{person}{Tamara~L Berg}.} \bibinfo{year}{2016}\natexlab{}.
\newblock \showarticletitle{Modeling context in referring expressions}. In
  \bibinfo{booktitle}{\emph{ECCV}}.
\newblock


\bibitem[\protect\citeauthoryear{Zellers, Bisk, Farhadi, and Choi}{Zellers
  et~al\mbox{.}}{2019}]%
        {zellers2019from}
\bibfield{author}{\bibinfo{person}{Rowan Zellers}, \bibinfo{person}{Yonatan
  Bisk}, \bibinfo{person}{Ali Farhadi}, {and} \bibinfo{person}{Yejin Choi}.}
  \bibinfo{year}{2019}\natexlab{}.
\newblock \showarticletitle{From Recognition to Cognition: Visual Commonsense
  Reasoning}. In \bibinfo{booktitle}{\emph{CVPR}}.
\newblock


\bibitem[\protect\citeauthoryear{Zhang, Li, Hu, Yang, Zhang, Wang, Choi, and
  Gao}{Zhang et~al\mbox{.}}{2021}]%
        {zhang2021vinvl}
\bibfield{author}{\bibinfo{person}{Pengchuan Zhang}, \bibinfo{person}{Xiujun
  Li}, \bibinfo{person}{Xiaowei Hu}, \bibinfo{person}{Jianwei Yang},
  \bibinfo{person}{Lei Zhang}, \bibinfo{person}{Lijuan Wang},
  \bibinfo{person}{Yejin Choi}, {and} \bibinfo{person}{Jianfeng Gao}.}
  \bibinfo{year}{2021}\natexlab{}.
\newblock \showarticletitle{Vinvl: Revisiting visual representations in
  vision-language models}. In \bibinfo{booktitle}{\emph{CVPR}}.
\newblock


\bibitem[\protect\citeauthoryear{Zhou, Palangi, Zhang, Hu, Corso, and Gao}{Zhou
  et~al\mbox{.}}{2020}]%
        {zhou2020unified}
\bibfield{author}{\bibinfo{person}{Luowei Zhou}, \bibinfo{person}{Hamid
  Palangi}, \bibinfo{person}{Lei Zhang}, \bibinfo{person}{Houdong Hu},
  \bibinfo{person}{Jason Corso}, {and} \bibinfo{person}{Jianfeng Gao}.}
  \bibinfo{year}{2020}\natexlab{}.
\newblock \showarticletitle{Unified vision-language pre-training for image
  captioning and vqa}. In \bibinfo{booktitle}{\emph{AAAI}}.
\newblock


\bibitem[\protect\citeauthoryear{Zhu, Zhao, Zhang, Ye, Chen, Zhang, and
  Chen}{Zhu et~al\mbox{.}}{2021}]%
        {zhu2021knowledge}
\bibfield{author}{\bibinfo{person}{Yushan Zhu}, \bibinfo{person}{Huaixiao
  Zhao}, \bibinfo{person}{Wen Zhang}, \bibinfo{person}{Ganqiang Ye},
  \bibinfo{person}{Hui Chen}, \bibinfo{person}{Ningyu Zhang}, {and}
  \bibinfo{person}{Huajun Chen}.} \bibinfo{year}{2021}\natexlab{}.
\newblock \showarticletitle{Knowledge perceived multi-modal pretraining in
  e-commerce}. In \bibinfo{booktitle}{\emph{ACM-MM}}.
\newblock


\bibitem[\protect\citeauthoryear{Zhuge, Gao, Fan, Jin, Chen, Zhou, Qiu, and
  Shao}{Zhuge et~al\mbox{.}}{2021}]%
        {zhuge2021kaleido}
\bibfield{author}{\bibinfo{person}{Mingchen Zhuge}, \bibinfo{person}{Dehong
  Gao}, \bibinfo{person}{Deng-Ping Fan}, \bibinfo{person}{Linbo Jin},
  \bibinfo{person}{Ben Chen}, \bibinfo{person}{Haoming Zhou},
  \bibinfo{person}{Minghui Qiu}, {and} \bibinfo{person}{Ling Shao}.}
  \bibinfo{year}{2021}\natexlab{}.
\newblock \showarticletitle{Kaleido-bert: Vision-language pre-training on
  fashion domain}. In \bibinfo{booktitle}{\emph{CVPR}}.
\newblock


\end{thebibliography}




\end{document}